%% file: main.tex
\pdfoutput=1
\documentclass[11pt]{article}

\usepackage[]{acl}

\usepackage{times}
\usepackage{latexsym}

\usepackage[T1]{fontenc}
\usepackage[utf8]{inputenc}

\usepackage{microtype}

\usepackage{inconsolata}

\usepackage{booktabs} %Yazeed, for figures
\usepackage{amsfonts} %Yazeed, for math notation
\usepackage{xcolor} %Yazeed, for table coloring
\usepackage{pifont}% Yazeed, for lit review table cross and check marks
\usepackage{algpseudocode}% Hisham, for pseudocode
\usepackage{algorithm}% Hisham, for pseudocode
\usepackage{amsmath} % Hisham, for pseudocode
\usepackage{enumitem}% Hisham, for lists
\newcommand{\cmark}{\ding{51}}%
\newcommand{\xmark}{\ding{55}}%
\usepackage{graphicx} %Yazeed, for figures
\usepackage{multirow} %Yazeed, for tables
\usepackage{cleveref}
\usepackage{makecell} % Hisham, for newlines in cells  

\title{When Benchmarks are Targets: Revealing the Sensitivity of Large Language Model Leaderboards}

\newcommand*\samethanks[1][\value{footnote}]{\footnotemark[#1]}

\author{Norah A. Alzahrani\thanks{Core contributor}, \textbf{Hisham Abdullah Alyahya}\samethanks, \textbf{Yazeed Alnumay}, \textbf{Sultan Alrashed}, \\
\textbf{Shaykhah Z. Alsubaie}, \textbf{Yousef Almushayqih}, \textbf{Faisal Abdulrahman Mirza}, \textbf{Nouf M. Alotaibi} \\
\textbf{Nora Al-Twairesh}, \textbf{Areeb Alowisheq}, \textbf{M Saiful Bari}, \textbf{Haidar Khan}\samethanks\thanks{Corresponding author: haidark@sdaia.gov.sa}
\\\\
National Center for AI (NCAI), Saudi Data and AI Authority (SDAIA) \\ Riyadh, Saudi Arabia
}

\newcommand{\hisham}[1]{\textcolor{brown}{(Hisham: #1)}}

\begin{document}
\maketitle

    \input{sections/00_abstract}
    \input{sections/10_introduction}
    \input{sections/30_methods}

\input{sections/35_experiments}
    \input{sections/40_results_and_analysis}

\input{sections/50_related_work}

\input{sections/99_conclusion}

    \input{sections/100_limitations}

% \input{sections/45_analysis}

\bibliography{custom}
\bibliographystyle{acl_natbib}
\pagebreak
\clearpage
\input{sections/101_appendix}
\end{document}

%% file: sections/00_abstract.tex
\begin{abstract}

Large Language Model (LLM) leaderboards based on benchmark rankings are regularly used to guide practitioners in model selection. Often, the published leaderboard rankings are taken at face value — we show this is a (potentially costly) mistake. Under existing leaderboards, the relative performance of LLMs is highly sensitive to (often minute) details. We show that for popular multiple-choice question benchmarks (e.g., MMLU), minor perturbations to the benchmark, such as changing the order of choices or the method of answer selection, result in changes in rankings up to 8 positions. We explain this phenomenon by conducting systematic experiments over three broad categories of benchmark perturbations and identifying the sources of this behavior. Our analysis results in several best-practice recommendations, including the advantage of a \textit{hybrid} scoring method for answer selection. Our study highlights the dangers of relying on simple benchmark evaluations and charts the path for more robust evaluation schemes on the existing benchmarks. The code for this paper is available at \href{https://github.com/National-Center-for-AI-Saudi-Arabia/lm-evaluation-harness}{https://github.com/National-Center-for-AI-Saudi-Arabia/lm-evaluation-harness}.
\end{abstract}

%% file: sections/10_introduction.tex
\section{Introduction}
\begin{figure*}[ht]
    \centering
    \includegraphics[width=\linewidth]{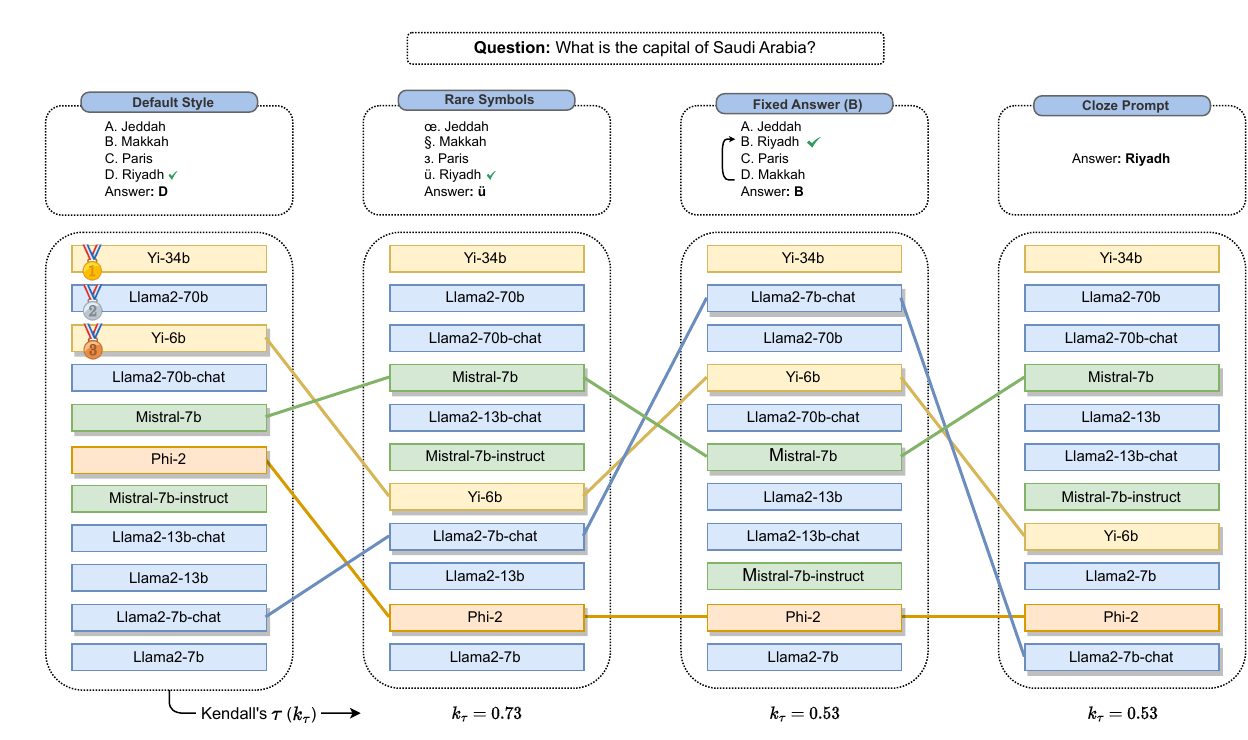}
    \caption{Minor perturbations cause major ranking shifts on MMLU~\cite{hendrycksmeasuring}. Models can move up or down up to eight positions on the leaderboard under small changes to the evaluation format. Columns (from left): 1) Original ranking given by MMLU using answer choice symbol scoring (a common default). 2) Ranking under an altered prompt for the same questions, where answer choice symbols are replaced with a set of rare symbols. 3) Setting where the correct answer choice is fixed to a certain position (in this case, B). 4) Using the cloze method for scoring answer choices. Under each new ranking, we report Kendall's $\tau$~\cite{kendall1938new} with respect to the original ranking (lower $k_\tau$ indicates more disagreement between rankings)}
    \label{fig:main_figure}
\end{figure*}

The advent of transformer-based Large Language Models (LLMs) \citep{openai2023gpt4,gemini,claude2,palm2,touvron2023llama2} has led to a generational leap in generative models, enabling interaction with computing devices through natural language. 
This advancement encompasses improvements that have rendered many earlier benchmarks and leaderboards obsolete \citep{laskar2023systematic,shen2023large}, leading to the compilation of more challenging and comprehensive tests. However, the current generation of leaderboards still does not satisfy many of the requirements of researchers and practitioners looking to build on LLMs~\cite{ethayarajh2021utility,dehghani2021benchmark}. Since LLMs are extremely expensive to both train and inference, selecting the LLM (or LLM training recipe) is often the most costly decision for the entire project. Stable leaderboards are critical to making the right decision.

% in \emph{fluency}, \emph{knowledge retrieval}, and \emph{truthfulness} that are then aimed to be aligned to human preferences \citep{christiano2017deep,stiennon2022learning,instructGPT}. 
% Additionally, the introduction of new tooling capabilities \footnote{\url{https://openai.com/blog/function-calling-and-other-api-updates}} \citep{schick2023toolformer,yang2023large,cai2023large,wong2023word} has presented challenges in tracking progress with traditional benchmarks. 
% This is primarily due to the perceived simplicity of evaluating LLMs on MCQs as well as the ease of compiling large and challenging MCQ benchmarks with diverse domains. 
% ~\Cref{tab:benchmark_popularity} illustrates the most commonly utilized benchmark in the latest LLM releases, with MMLU \citep{hendrycksmeasuring} – a multiple-choice QA benchmark – consistently chosen for assessing LLM capabilities.

Leaderboards based on multiple choice questions (MCQ) for evaluation~\citep{wang2018glue,super_glue,nie2019adversarial,zhong2023agieval,hendrycksmeasuring} present both convenience and significant limitations \citep{pezeshkpour2023large,zheng2023large}. While MCQs offer an \emph{automated} and \emph{quantifiable} means to assess certain aspects of model ability (e.g., knowledge), they fall short as a stable means to measure performance. Figure~\ref{fig:main_figure} demonstrates the instability of the leaderboard ranking of one popular benchmark, Massive Multitask Language Understanding (MMLU)~\cite{hendrycksmeasuring}, under small perturbations.

% capturing the full spectrum of a model's capabilities. The nuanced intricacies of \emph{(i)} language understanding, \emph{(ii)} contextual interpretation, \emph{(iii)} reasoning \cite{wei2022inverse,mckenzie2022inverse}, and \emph{(iv)} ethical consideration of the ability to handle generative scenarios, remain largely untested in this format.

Moreover, the reliance on MCQs raises concerns about the models being \emph{overfit} to these benchmarks, potentially excelling in structured tests while lacking real-world applicability. This discrepancy highlights the need for more holistic and diverse evaluation methods that transcend the simplicity of MCQs~\cite{liang2023holistic}. It also prompts critical reflection on how these models might inadvertently be trained to achieve high scores through spurious correlations, pattern recognition, and optimization for specific question formats rather than genuine language comprehension or knowledge. As LLMs continue to evolve, it is imperative to develop evaluation frameworks that can more accurately assess their abilities in a way that mirrors the complexity of real-world use. 

Despite being widely used, benchmarking with MCQs has turned out to be anything but simple. It requires the full synchronization of evaluation frameworks and results often vary wildly due to nuanced differences. For example, minor changes in prompting and scoring can produce invalid results for particular LLMs\footnote{\url{https://huggingface.co/blog/evaluating-mmlu-leaderboard}}. Recent studies have investigated the issue of sensitivity in evaluating LLMs, with some demonstrating that LLMs are susceptible to the ordering of answer choices and bias towards specific tokens/symbols~\citep{zheng2023large,pezeshkpour2023large,lu-etal-2022-fantastically}, and others exploring different prompt modifications and their effects in benchmarking LLMs~\citep{mizrahi2024state,sclar2023quantifying,weber-etal-2023-mind,pmlr-v139-zhao21c}.

In this work, we conduct a broad range of minor perturbation experiments to MCQ benchmarks and observe the disruption it causes to model rankings on leaderboards. We also take additional steps to precisely identify the limitations of LLMs on this measurement approach.

The contributions of this paper can be summarized as follows:

\begin{enumerate}
    \item Existing model rankings on popular benchmarks \textbf{break down under slight perturbations}, particularly in the medium to small model sizes.
    \item This behavior can be explained by the susceptibility of all tested LLMs to various forms of bias in MCQ. 
    \item Some families of LLMs have an over-reliance on format, pointing to potential benchmark leakage.
    \item We find that LLMs also exhibit bias to the scoring method for answer choices in MCQ.
    \item We demonstrate that some categories of modifications do not affect the benchmark rankings.

\end{enumerate}

% The rest of this paper is organized as follows: In the next section, we introduce the necessary background on LLM evaluation and the broad categories of tests we conduct. Section~\ref{methods} breaks down each category of test and details the setup. In Section~\ref{experiments} and Section~\ref{results} we specify experimental details and analyze the results respectively. Finally, we cover relevant prior work in Section~\ref{related} and conclude in Section~\ref{conclusion}

%% file: sections/30_methods.tex
\section{LLM Evaluation with MCQs}
\label{sec:LLM_Evaluation_with_MCQ}

Evaluating LLMs with MCQs has rapidly become a standard for measuring the knowledge and reasoning capabilities of the model \cite{openai2023gpt4,palm2,gemini,jiang2023mistral}. Many such MCQ benchmarks have been used to measure LLMs, including Massive Multitask Language Understanding (MMLU) \citep{hendrycksmeasuring}, Ai2 Reasoning Challenge (ARC) \citep{Clark2018ThinkYH}, and Common-sense Question Answers (CSQA) \citep{saha2018complex}. 

Mechanically, testing LLMs with MCQs is accomplished by presenting the question along with the answer choices to the model and selecting the choice deemed most probable by the model. 
% The method for choosing the answer varies and can be one of the following: \emph{(i)} Generating the token corresponding to the answer choice (e.g. ``C'' to select the third choice). \emph{(ii)} Comparing the likelihoods assigned to the token corresponding to each answer choice. \emph{(iii)} Comparing the normalized likelihoods of the content of each answer.
Although this setup appears straightforward, LLMs react in unpredictable ways to formatting and other minor changes to the questions or the answers. LLM performance on an MCQ test can change with the introduction of an extra space (e.g., between the question and answer) or adding an additional instructional phrase (e.g., "Choices:"). In addition to this brittleness, \citet{pezeshkpour2023large} found changes to the order in which answer choices are presented to GPT4 and instructGPT can change the model's prediction.

These findings lead us to take a deeper look at how MCQ-based benchmark results are affected by small perturbations to question formats, LLM prompts, presentation of few-shot examples, and other dimensions. In particular, we introduce variations in three categories:
\begin{itemize}
    \item \textbf{Answer choice format and ordering}: testing the limits of LLM sensitivity to ordering and formatting (Section~\ref{sec:choices}).
    \item \textbf{Prompt and scoring modifications}: changing text included in the prompt and analyzing different scoring schemes (Section~\ref{sec:Prompt_and_scoring_modifications}).
    \item \textbf{In-context knowledge manipulation}: inserting relevant/irrelevant information in the prompt and/or few-shot examples (Section~\ref{sec:incontext}).
\end{itemize}

Our main aim is to quantify how these small perturbations/variations \textbf{change the rankings} of a set of models on a particular benchmark. As MCQ benchmarks-based leaderboards are often used to compare models and guide model selection, we investigate the robustness of benchmarks for this purpose. Figure~\ref{fig:main_figure} demonstrates how existing benchmarks exhibit significant undesirable shifts in rankings under small perturbations.

\section{Methods}
\label{methods}
In this section, we describe and justify the perturbations we apply in each category. We note that some MCQ test changes, like modifying the order of answer choices, can change performance even for humans but the effect is typically not pronounced~\citep{lions2021position}. In general, our modifications are designed to be small perturbations to the MCQ and prompts that \emph{should not} affect performance. The exception to this is some of the \textbf{in-context knowledge manipulations} described in Section~\ref{sec:incontext}, which are designed to improve or degrade performance drastically.

\subsection{Answer choice format and ordering}
\label{sec:choices}
In light of earlier findings related to selection bias~\cite{zheng2023large, pezeshkpour2023large,lu-etal-2022-fantastically}, we investigate the effects of changes to the presented order of answer choices and changes to the symbols associated with answer choices.

\paragraph{Random choice order}
Our first study aims to uncover how dependent MCQ benchmark performance and rankings are on the original ordering of the answer choices.
We apply two simple schemes to randomly change the order of answer choices presented to the model: (i) swapping choices using a fixed set of swaps for all questions and (ii) randomly assigning new positions to each choice while ensuring each choice is moved to a different position.

\paragraph{Biased choice order}
In this setting, the correct answer choice is set to a fixed position across the entire test to measure bias toward predicting answers at particular positions. For zero-shot, we simply set the correct answer choice to each of the positions in turn.

\begin{figure}[ht]
    \centering
    \includegraphics[width=\linewidth]{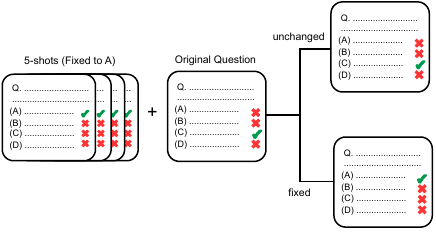}
    \caption{Experiment setup for probing position bias with few-shot examples.}
    \label{fig:fixed_few_shot_bias}
\end{figure}

In the few-shot case, we examine the influence of biasing the correct answers in the examples to the model's inherent bias to particular positions. For each question, we fix the correct answer of the examples to each position in turn. We then modify the test question in two ways: (i) unchanged answer choices and (ii) correct choice fixed to the same position as examples. This setup is shown in Figure~\ref{fig:fixed_few_shot_bias}.

\paragraph{Answer choice symbols}
\label{sec:Choice_ID_symbols}
The symbols used for the answer choices (e.g. A, B, C, D) also play a role in model bias~\cite{zheng2023large}. Thus, we experiment with replacing the symbols with alternative and less common tokens. This aims to decouple the bias to particular positions from the bias to symbols or the relative ordering in natural symbols. We replace \textit{['A', 'B',' C', 'D']} with the following two sets of symbols: 
\begin{itemize}
    \item Set 1: \textit{["\$", "\&", "\#", "@"]}  comprising of common tokens that are language-independent.
    % \item Set 2: ["œ", "§", "з", "ü"], they are unfamiliar characters that is from different alphabets sets which are unordered.
    \item Set 2: \textit{["œ", "§", "Ze (Cyrillic)", "ü"]} consisting of rare tokens in the vocabulary without any implicit relative order.
\end{itemize}

In the few-shot setting, we test both assigning fixed ordering for the replaced symbols in the examples as well as changing the ordering across examples.

% \begin{figure}[ht]
%     \centering
%     \includegraphics[width=0.85\linewidth]{figures/sample_of_alphabet_bias.drawio.pdf}
%     \caption{This figure illustrates a sample where all choices start with letter "B" and both llama-2-7b and llama-2-7b-chat logs (logs(a,b)) had the highest probability of "B" while the true was "D"}
%     \label{fig:sample_of_alphabet_bias}
% \end{figure}

\subsection{Prompt and scoring modifications}
\label{sec:Prompt_and_scoring_modifications}
LLMs exhibit high sensitivity to variations in prompt formatting \cite{T0, mishra2022crosstask}, forcing benchmark developers to unify prompt templates within the same evaluation scheme. However, it remains unknown if certain models have an affinity towards any specific prompt templating style. It is unclear how benchmarking prompt choices advantage/disadvantage different models. In addition to that, the scoring style may change depending on how we are prompting the context of a query. We distinguish three major categories of scoring methods for MCQs.

\begin{itemize}
    \item \textbf{Symbol scoring}\label{par:symbol}: Prompt template is structured as question followed by answer choices. The model chooses the answer based on the likelihood scores for the answer choice symbol. Used in \citet{hendrycksmeasuring}.
    \item \textbf{Hybrid scoring}: Prompt template is structured as a question followed by answer choices. The model chooses the answer based on the likelihood scores for the answer choice content normalized by length. Used in \citet{t5,T0,chowdhery2022palm}
    \item \textbf{Cloze scoring}: Prompt templates are structured as a question followed by a single answer choice. Maximum normalized likelihood scores over all answer choices define the prediction. Used in \citet{Clark2018ThinkYH}.
    
\end{itemize}

\begin{figure}[ht]
    % https://drive.google.com/file/d/1Wr5w4lHqF1jWmFp2lJm8C-Ub-8Syx6Tv/view?usp=sharing
    \centering
    \includegraphics[width=\linewidth]{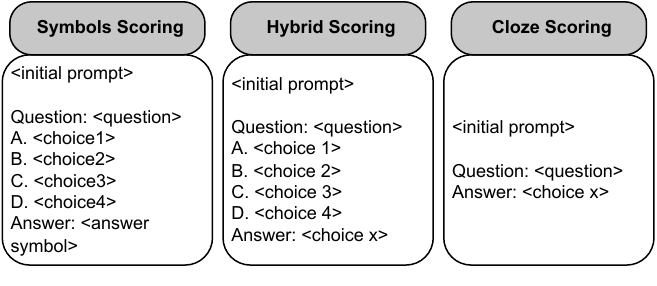}
    \caption{Answer choice scoring methods for LLMs. The symbols and hybrid scoring methods are most similar, sharing identical prompts. Cloze scoring does not reflect a “true” MCQ style, as the model is not shown all the options. However, due to its prevalence we compare it to the other methods as a baseline.}
    \label{fig:prompt_styling_and_scoring}
\end{figure}

Figure~\ref{fig:prompt_styling_and_scoring} gives an overview of each scoring method. In addition, we also investigate further modification of instruction and sentinel tokens in the prompt template. 

\paragraph{Prompt instructions}
To assess the impact of subtle token alterations in prompt instructions, we conduct experiments on \emph{(i)} removing question subject information and \emph{(ii)} adding "Correct" alongside the answer. These targeted changes aim to identify the robustness in response to certain tokens, particularly when they carry crucial information, as well as to evaluate the influence of contextual bias introduced by minor modifications of the instruction text. 
% These changes are illustrated in Figure~\ref{fig:varying_inst_prompts}.

%https://drive.google.com/file/d/1UiUUDfXdE1PzNkJEUrl5SI6-NVYscYj1/view?usp=sharing
% \begin{figure}[ht]
%     \centering
%     \includegraphics[width=0.85\linewidth]{figures/Varying_Instructional_formats_prompt_structures.drawio.pdf}
%     \caption{\color{purple}Illustration of each experiment mutation. Experiment 1 showcases the removal of the subject name from the instruction. Experiment 2 displays the modified instruction for zero-shot settings. Experiment 3 shows the prompt change by specifying 'Correct Answer' instead of 'Answer'.}
%     \label{fig:varying_inst_prompts}
% \end{figure}

% \paragraph{Sentinel tokens}
% LLMs are often finetuned with specific tags to easily parse structured data (i.e., multi-turn conversational data, system prompt, tabular data). The popular benchmarks \citep{gao2021framework,liang2023holistic} do not include these tokens into consideration while evaluating LLMs. Thus we investigate the effect of adding these model specific tags in the evaluation benchmark. For example, In \texttt{Llama-2-Chat}, \emph{instruction} and \emph{system prompts} are delimited by special sentinel tokens (i.e., \texttt{[INST]}, \texttt{[/INST]}, \texttt{<<SYS>>} and \texttt{<</SYS>>}) in a specific format.\footnote{\url{https://github.com/facebookresearch/llama/blob/main/example_chat_completion.py}}

\subsection{In-context knowledge manipulation}
\label{sec:incontext}
In this category, we attempt to measure model and benchmark robustness in the few-shot setting by testing the entire spectrum of knowledge injected in the few-shot examples. In particular, we experiment under the following settings:

\paragraph{Correct answer provided}
We provide the target question and the correct answer in the prompt as an example to the model. This corresponds to the simplest setting for the model, where it only needs to look up the answer in the context.

\paragraph{Incorrect answer provided}
This setting is the opposite of the former. The target question is provided with an incorrect answer as an example. It is challenging as the model must ignore the context and determine the correct answer independently.

\paragraph{Trivial examples}
We replace few-shot examples with simple questions the model is known to be able to answer (typically related to the language/text of the question itself). The only information the examples convey is related to formatting~\cite{soltan2022alexatm}. We create three versions of these questions and answers using GPT-4 and ensure the model can answer them correctly (as shown in Figure~\ref{fig:trivial_examples}).

%https://drive.google.com/file/d/11KmSHYUv7t9vbYCJuJoRM2qpRTNSXqNi/view?usp=sharing
% \begin{figure}[ht]
%     \centering
%     \includegraphics[width=0.85\linewidth]{figures/trivial_examples.drawio.pdf}
%     \caption{Illustration of the three versions of the trivial examples.}
%     \label{fig:trivial_examples}
% \end{figure}

\paragraph{Out of domain examples}
Instead of providing examples from the same subject as the target question, we add out-of-domain questions (from another subject) as the few-shot examples. This setting corresponds to a difficulty level between the original format and providing trivial examples.

%% file: sections/35_experiments.tex
\section{Experiments}
\label{experiments}
% To identify the most prevalent MCQ benchmarks, we collected the published benchmark results of popular open-source large language models, including Llama 2, Mistral, and Yi (Table \ref{tab:benchmark_popularity}). 

In the bulk of our experiments, we focus on the MMLU benchmark due to the extensive nature of our experiments (11 models, 22+ settings), and extend some experiments to ARC-challenge to show generalizability.
% We select three MCQ benchmarks for our study, MMLU, ARC-challenge, and ARC-easy.

MMLU~\cite{hendrycksmeasuring} is a commonly used benchmark for comparing LLMs, consisting of 57 subjects spanning four domains: humanities, STEM, social sciences, and others. Each subject includes at least 100 multiple-choice questions with 4 answer choices. The entire benchmark contains 14,042 questions (Tables ~\ref{tab:mmlu_test_stats} and ~\ref{tab:mmlu_dev_stats} have a breakdown of the MMLU subjects and their distributions).
% a sample question from the relatively easy elementary mathematics subject is presented in Figure \ref{fig:mmlu_format}. 

Ai2 Reasoning Challenge \cite{clark2018thinkarc} is a benchmark consisting of 7787 grade school science questions. The benchmark is split into two sets: Easy and Challenge. We conduct experiments on the Challenge set (ARC-C) which is proven to contain harder questions for existing models. The questions in ARC-C have 3-5 answer choices.

Unless otherwise stated, the reported score for each experiment/model combination on MMLU is the mean accuracy across all 14,042 questions. All tested model tokenizers encode the multiple-choice answers as single tokens. Hence, the accuracy is equivalent to the normalized accuracy. All baseline and modified MMLU benchmarks were performed using the LM Evaluation Harness \cite{eval-harness} library. Their implementation of MMLU measures the log-likelihood of each of the answer tokens \textit{['A', 'B', 'C', 'D']} after the input prompt and chooses the letter with the highest probability as the model's answer.

Some of our experiments require permuting the answer choice order (Table~\ref{tab:selected_domains_rand},~\ref{tab:selected_domain_rand_sym_content},~\ref{tab:selected_domains_rand_symb_options}, and ~\ref{tab:selected_domains_rand_style3}), however, this can be confusing for questions where the answer choices are dependent on their position, such as \textit{“D. All of the above.”}, or reference other choices, such as \textit{“C. Both A and B.”}. To circumvent this dependency, we manually inspected and modified the questions from three subjects to ensure their answers are permutation-independent for a subset of our experiments. The modified subjects are college chemistry, college mathematics, and global facts.

For each variation introduced to the MCQ benchmarks, we calculate the change in accuracy ($\Delta$Acc) and recall standard deviation (RStd) for each model. RStd measures the bias of a model to a particular answer choice by computing the standard deviation of recalls for each answer choice~\cite{zheng2023large}. This metric quantifies how much the model favors particular positions for the correct answer choice. We typically observe whether RStd changes ($\Delta$RStd) are significant across experimental settings. 

To measure the change in ranking induced by an applied perturbation to a benchmark, we measure the normalized Kendall's $\tau$ distance between two rankings of $n$ models~\cite{kendall1938new}. Kendall's $\tau$ computes the number of swapped pairs between two rankings normalized by the total number of pairs $\frac{n(n-1)}{2}$. We report $k_\tau = \frac{\tau+1}{2}$, where $k_\tau = 1.0$ indicates total agreement between rankings, and $k_\tau = 0$ indicates complete disagreement by reversing the original rankings.

%% file: sections/40_results_and_analysis.tex
\section{Results \& Analysis}
\label{results}
In this section, we highlight the major findings of our work and combine the results of multiple lines of experimentation (detailed in Section~\ref{methods}) into concise observations. Additional observations and complete experimental results can be found in the appendix (Section~\ref{sec:appendix}).
\subsection{MCQ benchmarks are not robust to perturbations}
As shown in Figure~\ref{fig:main_figure}, there exist perturbations that cause dramatic shifts in the order of models with respect to commonly accepted leaderboard rankings. We find a significant number of small perturbations demonstrate this effect, while other perturbations are more benign.
\paragraph{Sensitive perturbations} 
\begin{figure}[ht]
    \centering
    \includegraphics[width=\linewidth]{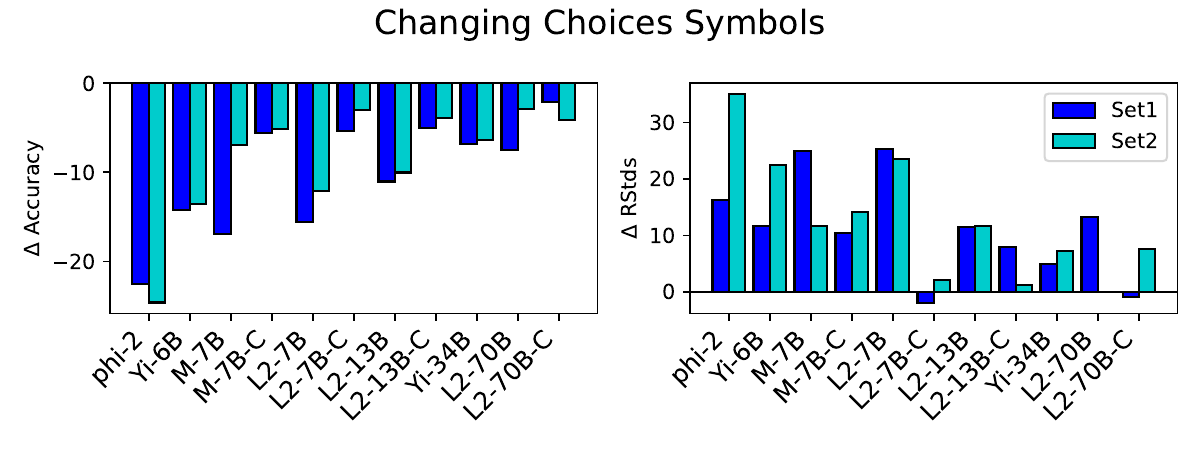}
    \caption{Change in accuracy and bias (RStd) on zero-shot MMLU after swapping answer choice symbols with two different sets of symbols (described in Section~\ref{sec:Choice_ID_symbols}). While accuracy always decreased, most models exhibited even more selection bias with the new symbols.
    $k_\tau$ for Set1 and Set2 were 0.689 and 0.733 respectively}
    \label{fig:mmlu_symbols_exp}
\end{figure}
Shuffling/changing the presented order of the choices, swapping choice symbols, and alternative scoring methods all cause major shifts to the rankings (determined by thresholding $k_\tau \le 0.75$). For example, in a controlled experiment using a subset of MMLU, we randomly shuffle the answer choices presented to the models (Table~\ref{tab:selected_domains_rand}). 5 out of 11 models change in ranking after the perturbation and $k_\tau$ drops to 0.564. A similar pattern is seen for perturbations like fixing the correct answer to a particular position (Table~\ref{tab:zero_shot_fixing}), replacing the default choice symbols with other sets (Figure~\ref{fig:mmlu_symbols_exp}), and alternative scoring methods (Figure~\ref{fig:diff_style}). 

Some models elicit this behavior much more strongly. For example, we observe that Yi-6b drops from 3rd place to 7th or 8th place under some benchmark perturbations in the group of 11 models we tested (namely, the rare symbol and cloze perturbations). Other models in the same size range are more stable (e.g., Mistral-7b, Llama2-7b), not shifting more than one or two ranks under all perturbations. The reasons for this are unclear but could indicate overfitting to aspects of the benchmark style. Since training data for these models is not public, it is difficult for us to verify this hypothesis.

\input{tables/selectedDomains_rand}
\paragraph{Unsensitive perturbations} Changes that have little effect on the model rankings are discussed in Section~\ref{sec:minor}.

\subsection{Revisiting selection bias: token bias vs. position bias}

Prior and concurrent work finds that LLMs answering MCQs are highly sensitive to the order that choices are presented~\cite{pezeshkpour2023large,robinson2023leveraging} (position bias) as well as the symbols used as choice IDs~\cite{zheng2023large} (token bias). We find selection bias is apparent in \textbf{all} LLMs we test both in 0 and 5-shot setups, as shown in Tables~\ref{tab:zero_shot_fixing} and~\ref{tab:five_shot_fixing}. This confirms earlier findings and highlights a major weakness of the current methods of evaluating LLMs on MCQs. 

\begin{figure}[ht]
    \centering
    \includegraphics[width=\linewidth]{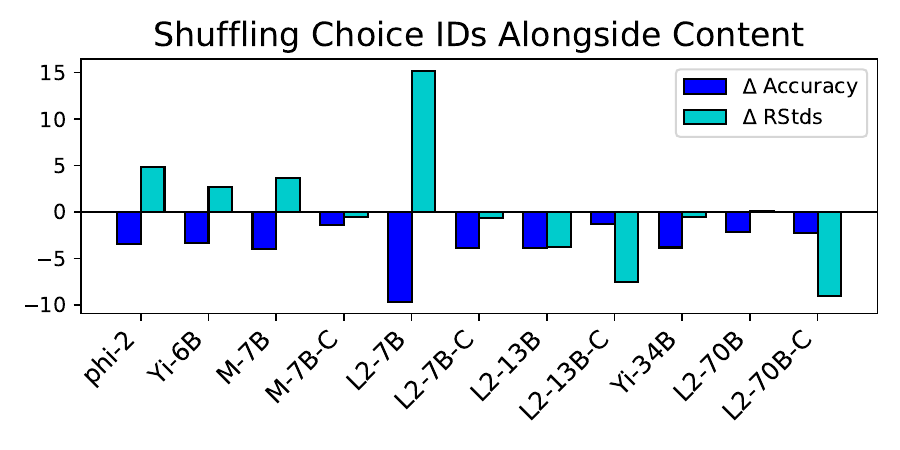}
    \caption{Accuracy and RStd change after randomly shuffling the order of the choices alongside their option IDs. Although \citep{zheng2023large} use this experiment as evidence that position bias has minimal effect on selection bias, we find it inconclusive as variance in $\Delta$RStd is large. 
    % $k_\tau$ = 0.956
    }
    \label{fig:shuffle_full_options}
\end{figure}

\begin{figure}[ht]
    \centering
    \includegraphics[width=\linewidth]{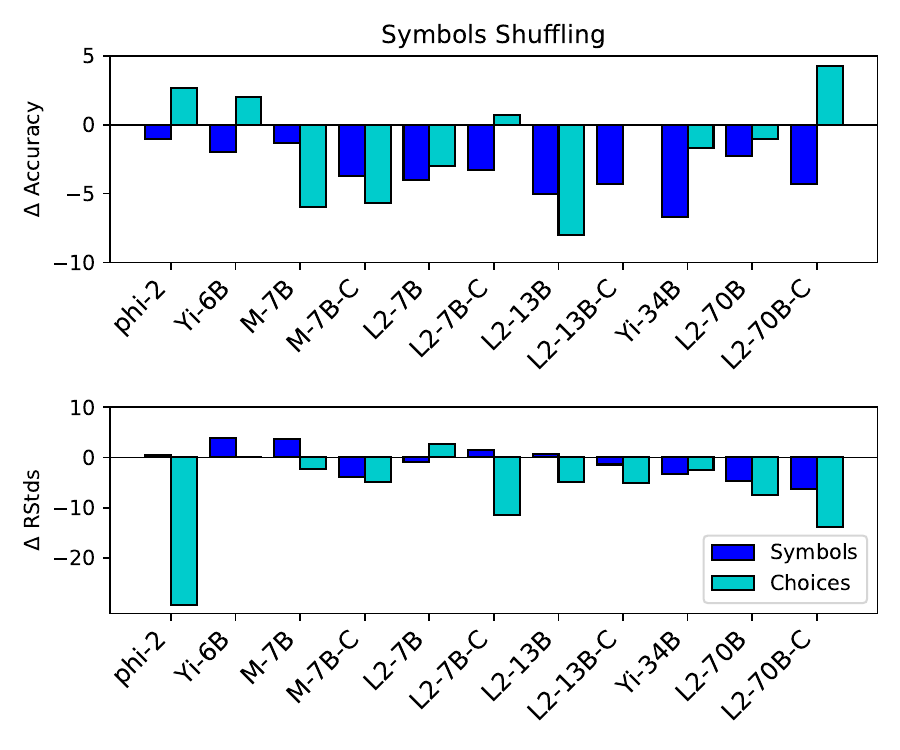}
    \caption{Using a set of rare symbols (Set2) we test two modes of shuffling answer choices: shuffling the symbols only (blue bars) and shuffling the answer choice text only while fixing the symbols set (cyan bars). Even using rare symbols, model selection bias changes unpredictably, indicating token and position bias are difficult to mitigate. This experiment was conducted using the three-subject subset of MMLU.}
    % The $k_\tau$ score for option shuffling is 0.467 while the $k_\tau$ score for choices shuffling is 0.733.
    
    \label{fig:symbols_rand}
\end{figure}

To disentangle these two sources of bias, we first measure the change in bias (measured by RStd) as we randomly shuffle the entire choice and symbols together, as performed in ~\cite{zheng2023large}. We find that simply shuffling entire choices is inconclusive in ruling out the effect of position bias (vs. token bias) as there is a wide variance in the bias change across LLMs (Figure~\ref{fig:shuffle_full_options}, Table~\ref{tab:shuffle_full_options} ). In light of this, we opt to isolate token bias from position bias by replacing the default symbols (A/B/C/D) with new/rare symbols from the LLM's vocabulary (without an implicit relative ordering) and shuffling them. This experiment, displayed in Figure~\ref{fig:symbols_rand} and Table~\ref{tab:selected_domain_rand_sym_content}, shows that (i) LLMs always bias toward the symbols representing the choice IDs and (ii) even after shuffling the symbols, bias changes in unpredictable ways.

\input{tables/zero_shot_option_fixing}
% \input{tables/shuffle_full_options}

% \input{tables/mmlu_symbols_set1_options_0shot}
% \input{tables/mmlu_symbols_set2_options_0shot}
% \subsubsection{Token Ablation}
% \hisham{
% The other ablation Tsinghua has run is to remove the option IDs entirely and finding that bias decreases for all models, arguing that it is sufficient evidence to conclude the primary cause of the bias is tied to the option IDs. We argue that the change in evaluation from token scoring to answer scoring was the primary reason behind this behavior, and not that the removal of the option IDs. [SHOW AND EXPLAIN ARC-STYLE RESULTS HERE]
% }
\subsection{Another source of bias: scoring bias}
\begin{figure}[ht]
    \centering
    \includegraphics[width=\linewidth]{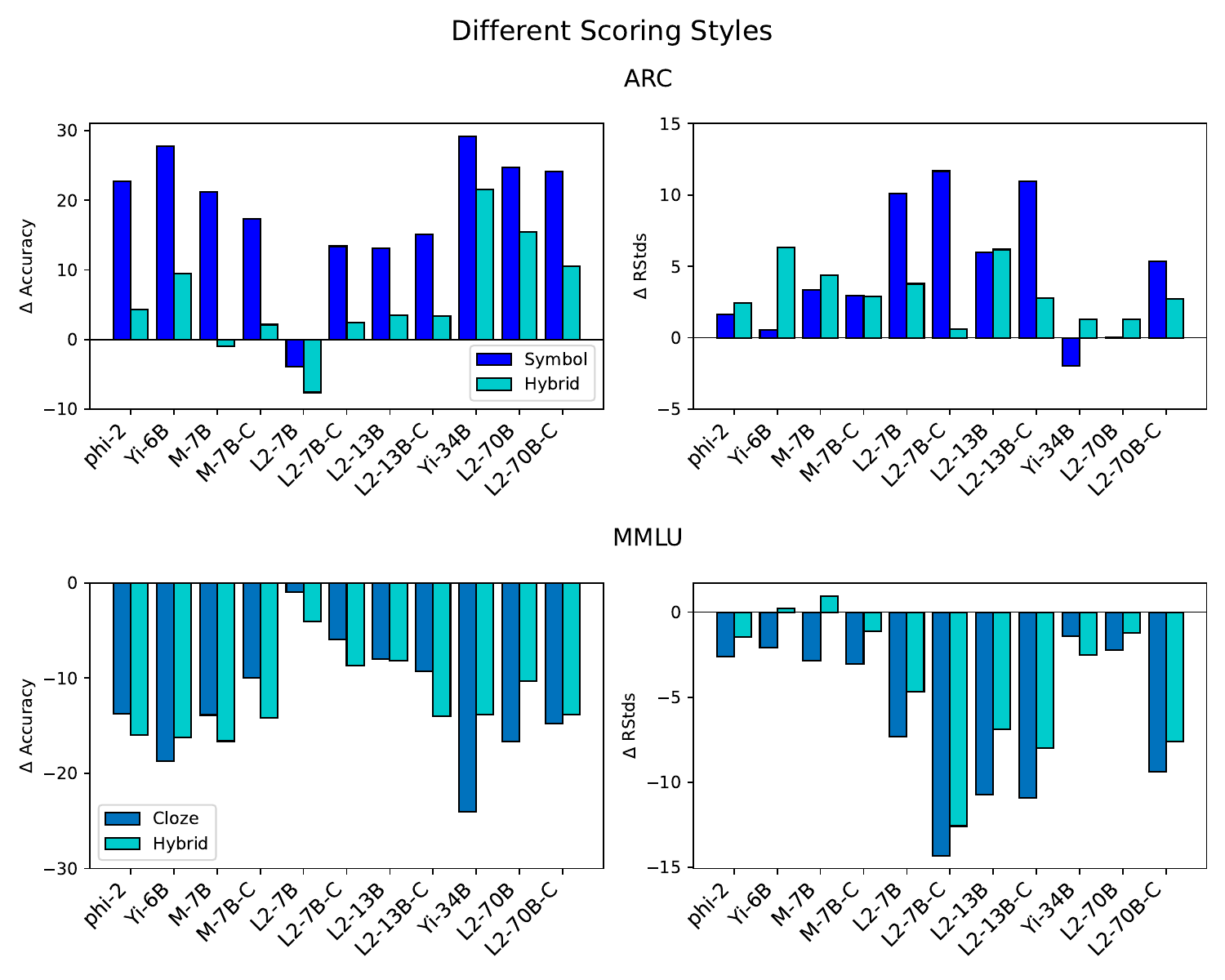}
    \caption{Comparing scoring method \textit{\{symbol, cloze, hybrid\}} across two tasks, MMLU and ARC-Challenge. Note the baseline method for MMLU is \textbf{symbol} while the baseline method for ARC-C is \textbf{cloze}. The general trend for accuracy across models and tasks is symbol scoring (highest accuracies) followed by hybrid scoring/cloze depending on the model. The measured selection bias also follows this trend, with symbol scoring resulting in the highest bias across models. 
    % $k_\tau$ scores for MMLU Cloze, MMLU Hybrid, ARC-C Symbols, and ARC-C Hybrid are  0.6, 0.778, 0.867, and 0.778, respectively.
    %The $k_\tau$ score for MMLU-Cloze = 0.6, MMLU-Hybrid = 0.778, ARC-C-Symbols = 0.867, ARC-C-Hybrid = 0.778
    }
    \label{fig:diff_style}
\end{figure}
Beyond the ordering of choices and the symbols associated with them, LLMs exhibit varying amounts of bias under the choice of scoring method for MCQs. We studied the three scoring methods described in Section~\ref{sec:Prompt_and_scoring_modifications}: symbol scoring, cloze scoring, and hybrid scoring. Symbol scoring has become the dominant method for evaluating LLMs on MCQs, largely due to the high accuracy achieved by LLMs~\cite{robinson2023leveraging}. This, however, comes at the cost of high selection bias. Cloze scoring can essentially eliminate bias since the choices are never presented to the model, but LLMs tend to score poorly when using this method. This also does not reflect a true MCQ setting. Figure~\ref{fig:diff_style} and Table~\ref{tab:mmlu_style2_choices_0shot} detail the results of these experiments. 

Hybrid scoring, where cloze scoring is combined with a prompt that reveals all answer choices to the model, represents an acceptable balance between the two, reducing bias over symbol scoring on MMLU and ARC-C, as shown in Figure~\ref{fig:diff_style}. In light of this, we recommend practitioners to replace symbol scoring with hybrid scoring to mitigate the effects of bias on model rankings.

% \begin{figure}[ht]
%     \centering
%     \includegraphics[width=0.90\linewidth]{figures/bais_final.PNG}
%     \caption{The average results of the shuffling experiment using the Synbols scoring styles compared with the Hybrid styles under zero-shot setting on the selected domains. Symbols scoring $k_\tau$ = 0.6 and Hybrid = 0.644}
%     \label{fig:rand_styles}
% \end{figure}

% \norah{To further investigate the Hybrid style, we shuffled the choices order as in Figure ~\ref{fig:rand_styles} and found that it also decreases the RStds values on average and had less drastic changes in the accuracy compared with the shuffling experiment using MCQs style. This led us to believe that the scoring style have a noticeable affect on the bias and it provides a more consistence results after shuffling while mostly reducing the RStds values. }

\subsection{Minor few-shot and prompt changes have little effect on benchmark rankings}
\label{sec:minor}

\begin{figure}[h!]
    \centering
    \includegraphics[width=0.85\linewidth]{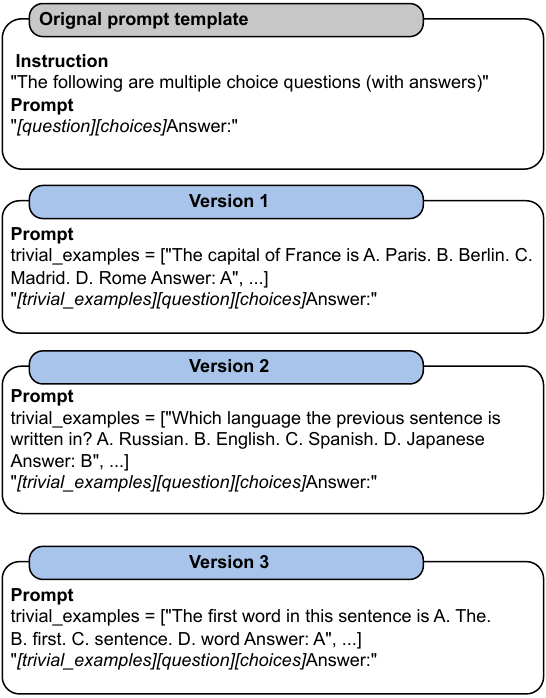}
    \caption{Illustration of the three versions of the trivial examples.}
    \label{fig:trivial_examples}
\end{figure}

\begin{figure}[h!]
    \centering
    \includegraphics[width=0.85\linewidth]{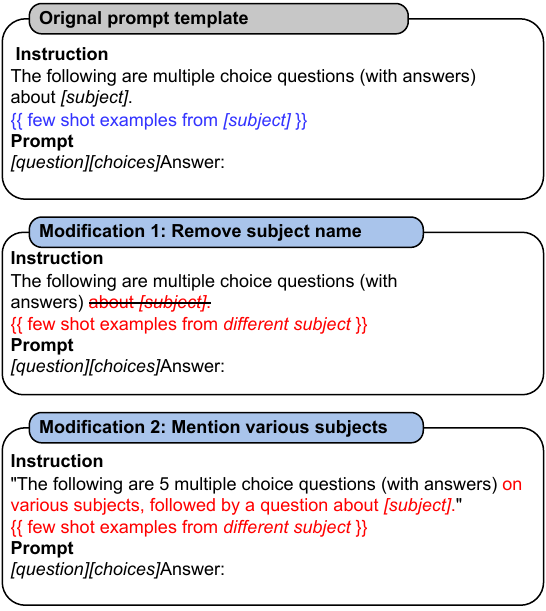}
    \caption{Illustration of subject independent few-shot prompting experiment. We ensure that we do not sample from similar domains to the one being evaluated (e.g. sampling college mathematics few-shots for high school mathematics questions). (results are in Table \ref{tab:subject_indep_fewshot1} \& \ref{tab:subject_indep_fewshot2}).}
    \label{fig:subject_independent_prompts}
\end{figure}

We ran several experiments to assess the effect of the initial prompt on model performance and rankings. We find that changing the informativeness of in-context examples, e.g. providing irrelevant/trivial examples (Figure~\ref{fig:trivial_examples}, Tables~\ref{tab:trivial_examples_exp1}-\ref{tab:trivial_examples_exp3}) or examples from subjects other than the target subject (Figure~\ref{fig:subject_independent_prompts}, Tables~\ref{tab:subject_indep_fewshot1}-~\ref{tab:subject_indep_fewshot2}), slightly changes performance across models and reduces bias compared to zero-shot settings but does not change rankings drastically.
This finding leads us to conclude that adding few shot examples to benchmark evaluations can help reduce, but not eliminate, leaderboard sensitivity.

% Both the trivial examples experiments and out-of-domain examples experiments, yielded a marginal enhancement in model performance compared to the 0-shot baseline, with out-of-domain examples slightly outperforming trivial ones. However, performance in both conditions fell short of the gains observed with 5-shot in-domain examples.

% Further analysis revealed that while in-context examples slightly increased the accuracy of base models more than fine-tuned models (on average 3\% compared to 1\% ), this increase had minimal impact on model rankings.
 
% Eliminating few-shot examples reduces evaluation costs, especially for larger models, .
% , and allow for a broader range of model capabilities to be assessed without additional expenses.

\begin{figure}[h!]
    \centering
    \includegraphics[width=0.9\linewidth]{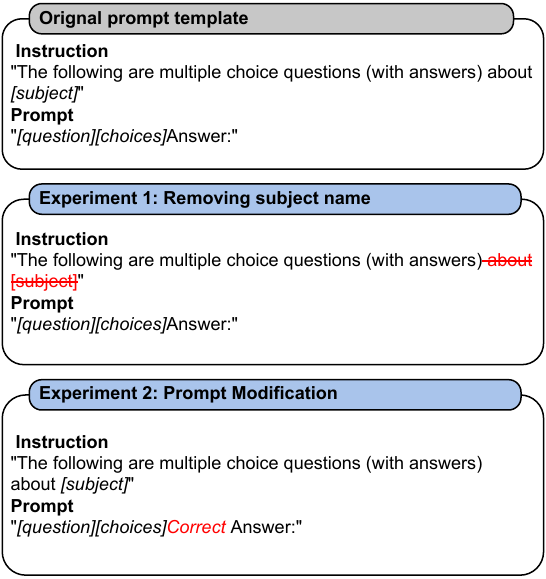}
    \caption{Illustration of minor prompt modifications. Experiment 1 showcases the removal of the subject name from the instruction. Experiment 2 shows the prompt change by specifying "Correct Answer" instead of "Answer". (results are in table \ref{tab:label_prompt_modification1}, \ref{tab:label_prompt_modification3}, \ref{tab:label_prompt_modification5})}
    \label{fig:varying_inst_prompts_shaykha}
\end{figure}

We also experiment with removing subject information from instructions and ending the prompt with "Correct Answer:" instead of "Answer:" (Figure~\ref{fig:varying_inst_prompts_shaykha}, Tables~\ref{tab:label_prompt_modification1}-\ref{tab:label_prompt_modification5}). We see little changes ($k_\tau > 0.9$) in these prompt modification experiments. 

\subsection{LLMs readily reference knowledge provided in-context (even if it is misleading)}
In our study of in-context knowledge injection, we find that LLMs can, expectedly, read off answers to questions when the answer is provided in the context (Table~\ref{tab:mmlu_cheating_combined}). However, when the question is answered incorrectly in the LLM's context (Table~\ref{tab:mmlu_cheating_wrong}), all models (regardless of size) are unable to reason correctly. 

To investigate whether this behavior is due to in-context knowledge acquisition or "incorrect answer" pattern following, previous work has shown that smaller models tend to rely more on priors learned during the pretraining stage while larger models tend to be more influenced by knowledge given in-context \cite{wei2023larger}. This result suggests that we would find that the magnitude of changes in accuracies for small models is smaller than that of larger models. However, we do not observe a significant effect to support that finding conclusively (when comparing \ref{tab:mmlu_baselines} with \ref{tab:mmlu_cheating_wrong}).

This behavior is further studied in \citet{wang2023resolving, xie2023adaptive,min-etal-2022-rethinking,yoo2022groundtruth} and indicates answer leakage in this way could affect benchmark results.

% When providing the models with correct answers in both 1-shot, and 5-shot settings, all models showed a drastic increase in accuracy, most reached above 0.90, with 5-shot results being mostly higher than 1-shot results. With the exception of both Llama 2-7b base and chat, although their accuracy increased in these experiments, but not as extremely as others. 

% The flipside of the above experiments was to provide the models with incorrect answers to assess the performance of various LLMs when faced with intentionally misleading prompts. Generally, we observed a large drop in accuracy across all models, a comparison between one-shot and five-shots settings for each model reveals that, in most cases, the one-shot setting yields slightly better performance compared to five-shot setting.

% Both settings show the susceptability of the models to in-context knowledge that contradicts their beliefs. This effect has been studied by ~\cite{wang2023resolving, xie2023adaptive}, which also show that LLMs beliefs tend to be affected by non-parametric knowledge and suggest approaches to mitigate the effect. Therefore, we recommend that this behavior should be studied more closely, and benchmarks created to rank models’ robustness to similar attacks as suggested by \citet{wang2023resolving}. Another note to make is that we also believe that this behavior is not attributed to sycophantic tendencies of aligned models, as this behavior appears in even base models.

To test whether LLMs can infer subtler patterns in the few-shots examples, we fix all answers in the few-shot examples to each of the positions A/B/C/D. The results (Table \ref{tab:five_shot_bias_unchanged}) suggest that LLMs also bias their answers to these kinds of (potentially inadvertent) patterns in the context.

While we have not observed these vulnerabilities in current benchmarks, we highlight them here as (potential) sources of benchmark instability.

\input{tables/5shot_bias_unchanged_short}
% \input{tables/cheatingWrong1shot}
% \input{tables/cheatingWrong5shots}
%Table~\ref{tab:cheating_wrong}
% \input{tables/trivial_examples_exp1}
% \input{tables/trivial_examples_exp2}
% \input{tables/trivial_examples_exp3}
% Table~\ref{tab:trivial_examples_exp1}

% \input{tables/fewshot_subject_indep}
%~\ref{tab:subject_indep_fewshot1, tab:subject_indep_fewshot2}. Table~\ref{tab:fake_fewshot}.
%Table~\ref{tab:five_shot_bias_unchanged} 
%\input{tables/5shot_bias_unchanged_short}

%% file: tables/selectedDomains_rand.tex
\begin{table}[ht]
    \centering
    \resizebox{\linewidth}{!}{%
        
            \begin{tabular}{lccc}%cc}
                \toprule
                \textbf{Model} & \textbf{Rank} & \textbf{Acc ($\Delta$Acc)} & \textbf{RStd ($\Delta$RStd)} \\
                \midrule
                \texttt{phi-2} & (7$\rightarrow$7) & 34.6 (-3) & 14.2 (7.4)\\% & 37.3 (-3.6)& 14.9(9.9)\\
                \texttt{Yi-6b} & (3$\rightarrow$9) & 33.0 (-8.3) & 11.9 (1.8) \\%& 35.667 (-5) & 12.894 (-1.2) \\
                \texttt{Mistral-7b} & (4$\rightarrow$3) & 40.0 (1.0) & 9.8 (0.7)\\% & 39 (-2) & 13.872 (1.8) \\
                \texttt{Mistral-7b-Instruct} & (8$\rightarrow$8) & 33.3 (-1.7) & 16.7 (3.5) \\% & 33.333 (-2.7) & 10.933 (-4.8) \\
                \texttt{Llama2-7b} & (11$\rightarrow$11) & 24.3 (-5.0) & 13.2 (-0.4) \\% & 29 (-4.3) & 22.634 (4.9) \\
                \texttt{Llama2-7b-chat} & (9$\rightarrow$10) & 28.6 (-3.7) & 27.7 (7.9) \\%& 28.667 (-4.7) & 17.837 (-3.5) \\
                \texttt{Llama2-13b} & (6$\rightarrow$6) & 37.0 (0.7) & 22.7 (5.7) \\%& 34.333 (-1.3) & 11.287 (-2.6) \\
                \texttt{Llama2-13b-chat} & (9$\rightarrow$5) & 37.6 (6.0) & 26.7 (0.0) \\%& 34.667 (2) & 18.168 (-6.5) \\
                \texttt{Yi-34b} & (1$\rightarrow$1) & 45.0 (-5.0) & 9.2 (-2.3) \\%& 48.667 (-0.7) & 8.759 (-0.6) \\
                \texttt{Llama2-70b} & (2$\rightarrow$2) & 40.3 (-1.7) & 9.07 (-5.5) \\%& 40 (-4.7) & 12.677 (6.5) \\
                \texttt{Llama2-70b-chat} & (5$\rightarrow$4) & 37.6 (0.3) & 13.4 (-6.2) \\%& 39.333 (-1.7) & 17.967 (-0.5) \\
                
%                 \midrule
% $k_\tau$ &  0.564 &\\%&0.709& \\
                \bottomrule
            \end{tabular}
        }

\caption{We show that model rankings can shift under shuffling of the order of answer choices. The largest change in rank is 5 positions (Yi-6b) followed by 4 positions (Llama2-13b-chat). This zero-shot experiment is done on a subset of MMLU subjects (college chemistry, college mathematics, and global facts) which we manually verified maintained correctness after shuffling answer choice order (i.e. did not contain cross references between answer choices). $k_\tau = 0.564$ for this experiment, indicating a significant disagreement in rankings.}
\label{tab:selected_domains_rand}
\end{table}

%% file: tables/zero_shot_option_fixing.tex
% makecell regex (\d+\.\d+) (\([+-]\d+\.\d+\)) -> \makecell{$1 \\\\ $2}
% ^\\texttt\{.*\} & (\d+\.\d+) & (\\make.*)

\begin{table}[ht]
\centering
\resizebox{\linewidth}{!}{
\begin{tabular}{lccccc}
\toprule
Model & \textbf{Baseline} & \textbf{A} & \textbf{B} & \textbf{C} & \textbf{D} \\
\midrule
\texttt{phi-2} & 54.47 & \makecell{52.31 \\ \textcolor{red}{(-2.16)}} & \makecell{56.53 \\ \textcolor{blue}{(+2.07)}} & \makecell{56.30 \\ \textcolor{blue}{(+1.83)}} & \makecell{50.19 \\ \textcolor{red}{(-4.28)}} \\
\texttt{Yi-6b} & 61.12 & \makecell{62.53 \\ \textcolor{blue}{(+1.41)}} & \makecell{64.44 \\ \textcolor{blue}{(+3.32)}} & \makecell{58.59 \\ \textcolor{red}{(-2.53)}} & \makecell{63.13 \\ \textcolor{blue}{(+2.02)}} \\
\texttt{Mistral-7b} & 59.56 & \makecell{52.19 \\ \textcolor{red}{(-7.38)}} & \makecell{60.98 \\ \textcolor{blue}{(+1.42)}} & \makecell{63.84 \\ \textcolor{blue}{(+4.27)}} & \makecell{60.43 \\ \textcolor{blue}{(+0.86)}} \\
\texttt{Mistral-7b-Instruct} & 53.48 & \makecell{49.77 \\ \textcolor{red}{(-3.71)}} & \makecell{54.67 \\ \textcolor{blue}{(+1.18)}} & \makecell{49.99 \\ \textcolor{red}{(-3.49)}} & \makecell{57.74 \\ \textcolor{blue}{(+4.26)}} \\
\texttt{Llama2-7b} & 41.81 & \makecell{66.36 \\ \textcolor{blue}{(+24.55)}} & \makecell{30.40 \\ \textcolor{red}{(-11.42)}} & \makecell{36.28 \\ \textcolor{red}{(-5.53)}} & \makecell{23.37 \\ \textcolor{red}{(-18.44)}} \\
\texttt{Llama2-7b-chat} & 46.37 & \makecell{30.84 \\ \textcolor{red}{(-15.53)}} & \makecell{69.41 \\ \textcolor{blue}{(+23.04)}} & \makecell{50.05 \\ \textcolor{blue}{(+3.68)}} & \makecell{28.23 \\ \textcolor{red}{(-18.14)}} \\
\texttt{Llama2-13b} & 52.08 & \makecell{35.82 \\ \textcolor{red}{(-16.26)}} & \makecell{57.24 \\ \textcolor{blue}{(+5.16)}} & \makecell{68.65 \\ \textcolor{blue}{(+16.57)}} & \makecell{44.08 \\ \textcolor{red}{(-8.00)}} \\
\texttt{Llama2-13b-chat} & 53.12 & \makecell{36.73 \\ \textcolor{red}{(-16.39)}} & \makecell{56.72 \\ \textcolor{blue}{(+3.60)}} & \makecell{71.81 \\ \textcolor{blue}{(+18.69)}} & \makecell{42.63 \\ \textcolor{red}{(-10.49)}} \\
\texttt{Yi-34b} & 73.38 & \makecell{66.16 \\ \textcolor{red}{(-7.22)}} & \makecell{75.22 \\ \textcolor{blue}{(+1.84)}} & \makecell{78.07 \\ \textcolor{blue}{(+4.69)}} & \makecell{73.88 \\ \textcolor{blue}{(+0.50)}} \\
\texttt{Llama2-70b} & 65.44 & \makecell{56.47 \\ \textcolor{red}{(-8.97)}} & \makecell{67.38 \\ \textcolor{blue}{(+1.95)}} & \makecell{69.92 \\ \textcolor{blue}{(+4.48)}} & \makecell{66.47 \\ \textcolor{blue}{(+1.03)}} \\
\texttt{Llama2-70b-chat} & 61.11 & \makecell{41.78 \\ \textcolor{red}{(-19.34)}} & \makecell{62.24 \\ \textcolor{blue}{(+1.13)}} & \makecell{75.07 \\ \textcolor{blue}{(+13.96)}} & \makecell{57.71 \\ \textcolor{red}{(-3.41)}} \\
\midrule
\text{$k_\tau$} & - & 0.455 & 0.527 & 0.527 & 0.855 \\
% \texttt{K_tau} & 61.110 & \makecell{41.775 \\ \textcolor{red}{(-19.335)}} & \makecell{62.242 \\ \textcolor{blue}{(+1.132)}} & \makecell{75.068 \\ \textcolor{blue}{(+13.958)}} & \makecell{57.705 \\ \textcolor{red}{(-3.405)}} \\
\bottomrule
\end{tabular}
}

\caption{
Performance on zero-shot MMLU when placing the correct answer at each possible position. All the LLMs tested showed a clear preference for specific positions/answer choice symbols, although the position varied among models and even in model families. These results corroborate the findings in \cite{zheng2023large}.
}
\label{tab:zero_shot_fixing}
\end{table}

%% file: tables/5shot_bias_unchanged_short.tex
% makecell regex (\d+\.\d+) (\([+-]\d+\.\d+\)) -> \makecell{$1 \\\\ $2}

\begin{table}[h]
\centering
\resizebox{\linewidth}{!}{
\begin{tabular}{lccccc}
\toprule
 & \textbf{5-shot Baseline} & \textbf{A} & \textbf{B} & \textbf{C} & \textbf{D} \\
\midrule
\texttt{phi-2} & 56.77 & \makecell{36.67 \\ \textcolor{red}{(-20.11)}} & \makecell{41.33 \\ \textcolor{red}{(-15.44)}} & \makecell{40.67 \\ \textcolor{red}{(-16.11)}} & \makecell{41.67 \\ \textcolor{red}{(-15.11)}} \\
\texttt{Yi-6B} & 63.22 & \makecell{36.67 \\ \textcolor{red}{(-26.56)}} & \makecell{36.33 \\ \textcolor{red}{(-26.89)}} & \makecell{37.67 \\ \textcolor{red}{(-25.56)}} & \makecell{39.33 \\ \textcolor{red}{(-23.89)}} \\
\texttt{Mistral-7B} & 62.36 & \makecell{34.67 \\ \textcolor{red}{(-27.70)}} & \makecell{41.33 \\ \textcolor{red}{(-21.03)}} & \makecell{43.00 \\ \textcolor{red}{(-19.36)}} & \makecell{40.33 \\ \textcolor{red}{(-22.03)}} \\
% Mistral-7B-Instruct-v0.1 & 53.952 & \makecell{32.667 \\ \textcolor{red}{(-21.285)}} & \makecell{33.333 \\ \textcolor{red}{(-20.619)}} & \makecell{30.667 \\ \textcolor{red}{(-23.285)}} & \makecell{35.333 \\ \textcolor{red}{(-18.619)}} \\
\texttt{Llama-2-7b} & 45.88 & \makecell{22.00 \\ \textcolor{red}{(-23.88)}} & \makecell{31.00 \\ \textcolor{red}{(-14.88)}} & \makecell{30.67 \\ \textcolor{red}{(-15.22)}} & \makecell{34.33 \\ \textcolor{red}{(-11.55)}} \\
\bottomrule
\end{tabular}
}

\caption{Results of fixing the 5 few-shot example answers to positions A/B/C/D on one model from each family, averaged over 3 selected subjects. We can see that performance drops across all cases/models, suggesting that models refer to subtle patterns in the context while answering. Full results are reported in Table \ref{tab:five_shot_bias_unchanged_long}}
\label{tab:five_shot_bias_unchanged}
\end{table}

%% file: sections/50_related_work.tex
\section{Related Work}
\label{related}

Benchmarks for the evaluation of LLMs \cite{chang2023survey} such as MMLU \cite{hendrycksmeasuring}, HELM \cite{liang2023holistic}, and BigBench \cite{suzgun2022challenging} have seen widespread adoption recently. Depending on the ability that is being assessed (e.g., language generation, knowledge understanding, complex reasoning) some benchmarks are designed in the form of close-ended problems like MCQs. 
To facilitate comparisons among LLMs, a number of leaderboards aggregating these benchmarks have been established, such as the OpenLLM Leaderboard  \cite{open-llm-leaderboard} and OpenCompass \cite{2023opencompass}. 
% Nonetheless, contemporary research \cite{zheng2023large,pezeshkpour2023large} reveals that multiple-choice questions (MCQs), as an evaluative tool for LLMs, lack robustness. 
% This deficiency is attributed to the intrinsic biases of LLMs, which may be oriented towards the identifiers of the answer choices or their positional arrangement within the answer set.

However, issues with the leaderboards and the underlying benchmarks have emerged. In a case study,  \citet{deng2023benchmark} discovered contamination/leakage of the MMLU benchmark in the training sets of multiple models. 
% They demonstrated their claim using two approaches: providing a partially masked benchmark question, and providing the question and answer choices with a single answer masked. In both cases, the model is prompted to fill the masks. 
A significant portion of models memorized benchmark questions and was able to perfectly reconstruct the removed part of some benchmark questions or answers. For instance, GPT-4 correctly completed the questions in 29\% of the prompts with URL hinting.

Even under the assumption of uncontaminated data, the performance of models on the underlying benchmarks are not robust to minor perturbations. \citet{pezeshkpour2023large} showed that specific orderings of MMLU answer choices resulted in up to $\pm 30\%$ deviations in GPT-4 performance on various subjects. Similarly, \citet{zheng2023large} demonstrate that models are biased to certain answer letters. On \texttt{llama-30B}, they showed a 27\% difference in MMLU accuracy by forcing all correct answers to either position \texttt{A} or \texttt{D}.
As well, \cite{robinson2023leveraging} find that the accuracy of LLMs improve (without regard to bias) when evaluating using a pure multiple choice question style vs a cloze question answering style.

While prior work has highlighted weaknesses in LLMs themselves~\cite{zheng2023large,pezeshkpour2023large}, evaluation method~\cite{robinson2023leveraging}, or the contents of benchmarks~\cite{dehghani2021benchmark} in our work we thoroughly study the effects these factors have on existing leaderboards and demonstrate where leaderboards lack robustness.

%% file: sections/99_conclusion.tex
\section{Conclusion}
\label{conclusion}

Building robust leaderboards is a major challenge for the community, as leaderboards help practitioners select the best methods and models for continued research. Given this importance, it is critical to address the breakdown of existing leaderboards to the slight perturbations we demonstrated in our work. In addition to building our understanding of the causes of this sensitivity (e.g., bias in LLMs and bias in scoring methods), future work should aim to adopt and design benchmark practices that avoid these pitfalls.

%% file: sections/100_limitations.tex
\section{Limitations}

The limitations of our work fall into two main categories: (i) understanding the causes of LLM bias and (ii) our limited success at overcoming leaderboard sensitivity.

To explain LLM bias, we attempted to design experiments that isolate each source of bias under MCQ but were unable to quantify the relative effects of bias or conclude why they occur. This was further complicated by our inability to access the pretraining datasets of the LLMs to rule out benchmark contamination. Future work in this direction will most likely require tools from interpretability research (e.g. mechanistic interpretability).

One of our main contributions was to highlight where MCQ-based leaderboards fail to deliver stable rankings. Although we succeeded in showing this, we were unable to demonstrate a robust solution to this problem. Our recommendation to, for example, use hybrid scoring methods is still not completely robust to perturbations.

% More broadly, our work’s objective was to highlight cases where MCQ benchmarks fail to align with our definition of performance~\cite{ethayarajh2021utility}. While we managed to show abnormalities in some experiments, there are potentially an infinite number of perturbations that could cause similar shifts. Our recommendation for using hybrid scoring will work until another perturbation is found that will prove it not to be robust. Because of this, it might be unwise to continue searching for these perturbations one by one. We think future research should attempt to investigate deeper issues in LLM benchmarking and find the root cause of this discrepancy between the score we value and the score valued by the benchmark.

\section{Potential Risks}
In this work, we do not present a new leaderboard. There is a risk that Figure~\ref{fig:main_figure} is interpreted as a leaderboard or used for model selection. Our intention was to demonstrate how minor changes can affect model rankings.

% \section{Acknowledgements}

% We thank Ehsan Hoque for valuable feedback on early versions of the manuscript.

%% file: sections/101_appendix.tex
\appendix

% prepend A. in front of sections, tables and figures in the appendix
\renewcommand{\thesection}{A.\arabic{section}}
\renewcommand{\thefigure}{A.\arabic{figure}}
\renewcommand{\thetable}{A.\arabic{table}}

% Reset the section, figure and table counter
\setcounter{section}{0}
\setcounter{figure}{0}
\setcounter{table}{0}

\section{Appendix}
We present a comprehensive collection of tables containing the results of all our experiments. The often complex nature of the observed behavior warrants a closer look that may inspire novel interpretations for future studies. We believe providing these detailed results will help researchers conduct further analysis and generate hypotheses to help drive research in LLM-benchmarking robustness forward.
\label{sec:appendix}

\subsection{Baselines}
This section lists the baselines referenced in different experiments throughout the paper.
\input{tables/mmlu_baselines}
\input{tables/arc_baselines}
\input{tables/selectedDomain_baseline}
\input{tables/selectedDomain_symbolsSet2_baseline}
\input{tables/selectedDomain_baseline_Style3}

\subsection{Answer choice format and ordering}
The following tables provide details on the choice formatting manipulation on the three selected MMLU subjects.
\input{tables/five_shot_option_fixing}
\input{tables/shuffle_full_options}
\input{tables/selectedDomains_randSymContent}
\input{tables/selectedDomains_randSymOptions}
\input{tables/selectedDomains_rand_style3}

\subsection{Prompt and scoring modifications}
The following tables provide results on the effect of different scoring styles of MCQs task on MMLU and ARC-C.

% \begin{figure}[H]
%     \centering
%     \includegraphics[width=0.85\linewidth]{figures/mmlu_symbols_options2.drawio.pdf}
%     \caption{Illustration of the changing of symbols on MMLU benchmark to Symbols Set1}
%     \label{fig:varying_symbols}
% \end{figure}

\input{tables/mmlu_symbols_set1_options_0shot}
\input{tables/mmlu_symbols_set2_options_0shot}

\input{tables/mmlu_style2_0shot}
\input{tables/mmlu_style3_0shot}
\input{tables/arc_c_0shot_Style1}
\input{tables/arc_c_0shot_Style3}

\subsection{In-context Knowledge Manipulation}
This section provides the results from experimentation on in-context manipulation.
%https://drive.google.com/file/d/11KmSHYUv7t9vbYCJuJoRM2qpRTNSXqNi/view?usp=sharing

\begin{figure}[h!]
    \centering
    \includegraphics[width=0.85\linewidth]{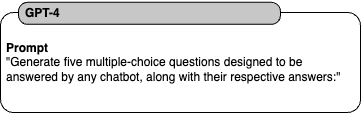}
    \caption{Illustration of the prompt that was used to generate the trivial examples version 1 using GPT4.}
    \label{fig:GPT-4_prompt_trivial_examples}
\end{figure}
\input{tables/trivial_examples_exp1}
\input{tables/trivial_examples_exp2}
\input{tables/trivial_examples_exp3}

\input{tables/varying_instructions_prompts.tex}

\input{tables/fewshot_subject_indep}

\input{tables/mmlu_cheating_wrong}
\input{tables/mmlu_cheating}

\input{tables/5shot_bias_unchanged_long}

\subsection{MMLU Overview}
\input{tables/mmlu_test_stats.tex}
\input{tables/mmlu_dev_stats.tex}

%% file: tables/mmlu_baselines.tex
\begin{table}[h!]
\centering
\resizebox{\linewidth}{!}{
% \begin{tabular}{lcccc}
% \hline
% Model & Acc 0shot & Acc 5shot \\ 
% \midrule

% \texttt{phi-2} & 54.47 &  56.77  \\
% \texttt{Yi-6B} & 61.12 & 63.23  \\
% \texttt{Mistral-7B} & 59.56 & 62.36  \\
% \texttt{Mistral-7B-Instruct} & 53.48 & 53.95  \\
% \texttt{Llama-2-7b} & 41.81 & 45.88  \\
% \texttt{Llama-2-7b-chat} & 46.37 & 47.22  \\
% \texttt{Llama-2-13b} & 52.08 & 55.06  \\
% \texttt{Llama-2-13b-chat} & 53.12 & 53.53  \\
% \texttt{Yi-34B} & 73.38 & 76.39  \\
% \texttt{Llama-2-70b} & 65.44 & 68.78  \\
% \texttt{Llama-2-70b-chat} & 61.11 & 63.17  \\

% % \midrule
% \bottomrule
% \end{tabular}
\begin{tabular}{lccccc}
\hline
Model & Acc 0shot & RStd 0shot & Acc 5shot & RStd 5shot \\ 
\midrule
\texttt{phi-2} & 54.47 & 4.01 & 56.77 & 2.65 \\
\texttt{Yi-6B} & 61.12 & 3.57 & 63.23 & 2.54 \\
\texttt{Mistral-7B} & 59.56 & 4.13 & 62.36 & 1.64 \\
\texttt{Mistral-7B-Instruct} & 53.48 & 4.58 & 53.95 & 4.78 \\
\texttt{Llama-2-7b} & 41.81 & 8.49 & 45.88 & 8.92 \\
\texttt{Llama-2-7b-chat} & 46.37 & 16.11 & 47.22 & 12.15 \\
\texttt{Llama-2-13b} & 52.08 & 12.04 & 55.06 & 4.42 \\
\texttt{Llama-2-13b-chat} & 53.12 & 12.80 & 53.53 & 8.32 \\
\texttt{Yi-34B} & 73.38 & 5.17 & 76.39 & 2.16 \\
\texttt{Llama-2-70b} & 65.44 & 3.20 & 68.78 & 1.56 \\
\texttt{Llama-2-70b-chat} & 61.11 & 10.95 & 63.17 & 8.06 \\
\bottomrule
\end{tabular}
}

\caption{The baseline accuracies and RStd values for the original MMLU implementation which uses the Symbols scoring style mentioned in section \ref{sec:Prompt_and_scoring_modifications}. All the models performed better in five-shot settings; the highest model was Yi-34B model in both settings. }
\label{tab:mmlu_baselines}
\end{table}

%% file: tables/arc_baselines.tex
\begin{table}[h!]
\centering
\resizebox{\linewidth}{!}{
\begin{tabular}{lcccc}
\hline
Model & Acc 0shot & RStd 0shot  &Acc 5shot &  RStd 5shot \\ 
\midrule

\texttt{phi-2} & 54.096 &2.558
& 58.874 & 2.509
 \\
\texttt{Yi-6B} & 50.512 & 2.114
 & 55.034 & 0.737
\\
\texttt{Mistral-7B} & 53.584 & 2.578
 & 59.556 &1.037
 \\
\texttt{Mistral-7B-Instruct} & 52.048 & 1.443
 & 54.778  &2.022
\\
\texttt{Llama-2-7b} & 46.331 & 4.094
 & 53.072 & 0.837
 \\
\texttt{Llama-2-7b-hf} & 44.283 & 2.175
 & 51.877 & 1.399
 \\
\texttt{Llama-2-13b} & 48.976 & 2.923
 & 56.997 & 0.799
 \\
\texttt{Llama-2-13b-chat} & 50.256 & 1.841
& 57.594 & 2.991
\\
\texttt{Yi-34B} & 61.519 & 2.537
 & 64.505 & 1.48
 \\
\texttt{Llama-2-70b} & 57.253 & 2.657
 & 66.126 & 1.926
 \\
\texttt{Llama-2-70b-chat} & 54.266 &1.505
 & 64.078 & 2.084
\\

\bottomrule
\end{tabular}
}

\caption{The baseline accuracies and RStd values for ARC-C using the Cloze scoring style mentioned in section \ref{sec:Prompt_and_scoring_modifications} which is considered as the original ARC-C implementation. As the table shows, the RStd values are relatively low in both settings. Yi-34B has the highest values on zero-shot while Llama-2-70b was the highest on five-shots }
\label{tab:arc_baselines}
\end{table}

%% file: tables/selectedDomain_baseline.tex
\begin{table}[ht!]
    \centering
    \resizebox{\linewidth}{!}{%
            \begin{tabular}{lcccc}
                \toprule
                Model&Acc 0shot&RStd 0shot&Acc 5shot&RStd 5shot \\
\midrule
\texttt{phi-2} & 37.67 & 6.78 & 41.00  & 5.02 \\
\texttt{Yi-6B} & 41.33 & 10.17 & 40.67 & 14.07 \\
\texttt{Mistral-7B} & 39.0 & 9.17 & 41.00 & 12.08 \\
\texttt{Mistral-7B-Instruct} & 35.0 & 13.31 & 36.00 & 15.75 \\
\texttt{Llama-2-7b} & 29.33 & 13.64 & 33.33 & 17.69 \\
\texttt{Llama-2-7b-chat} & 32.33 & 19.83 & 33.33 & 21.39 \\
\texttt{Llama-2-13b} & 36.33 & 17.05 & 35.67 & 13.85 \\
\texttt{Llama-2-13b-chat} & 31.67 & 26.78 & 32.67 & 24.69 \\
\texttt{Yi-34B} & 50.00 & 11.49 & 49.33 & 9.35 \\
\texttt{Llama-2-70b} & 42.00 & 14.58 & 44.67 & 6.21 \\
\texttt{Llama-2-70b-chat} & 37.33 & 19.63 & 41.00 & 18.46 \\
                \bottomrule
            \end{tabular}
        }

\caption{The selected three domains baseline average results on zero-shot and five-shot using Symbols scoring style on MMLU. MMLU mostly uses this scoring style. This baseline was utilized in most experiments to analyze and comprehend the influence of each experiment compared with this baseline in the selected domains subset (it was used in \ref{tab:selected_domains_sym_set2}, \ref{tab:selected_domains_baseline_style3} and \ref{tab:selected_domains_rand}).}
\label{tab:selected_domains_baseline}
\end{table}

%% file: tables/selectedDomain_symbolsSet2_baseline.tex
\begin{table}[h!]
    \centering
    \resizebox{\linewidth}{!}{%
            \begin{tabular}{lcc}
                \toprule
                Model & Task Acc ($\Delta$Acc) & Task RStd ($\Delta$RStd) \\
\midrule
\texttt{phi-2} & 26.33(-11.3) & 41.85 (35.0)\\ 
\texttt{Yi-6B} & 32.60 (-8.7) & 22.80 (12.7) \\
\texttt{Mistral-7B} & 35.30 (-3.7) & 18.79 (9.6) \\
\texttt{Mistral-7B-Instruct} & 34.00 (-1.0) & 26.90 (13.7) \\
\texttt{Llama-2-7b} & 29.60 (0.3) & 25.80 (12.2) \\
\texttt{Llama-2-7b-chat} & 31.30 (-1.0) & 27.00 (7.2) \\
\texttt{Llama-2-13b} & 34.30 (-2.0) & 26.10 (9.1) \\
\texttt{Llama-2-13b-chat} & 34.00 (2.3) & 21.90 (-4.8) \\
\texttt{Yi-34B} & 42.60 (-7.3) & 22.7 (11.3) \\
\texttt{Llama-2-70b} & 39.60 (-2.3) & 15.10 (0.5) \\
\texttt{Llama-2-70b-chat} & 36.00 (-1.3) & 29.50 (10.0) \\

                \midrule
$k_\tau$ = 0.527 \\
                \bottomrule
            \end{tabular}
       }
    
\caption{The baseline average zero-shot results for the selected domains using symbols Set2 which replaced the A/B/C/D choices symbols with œ/§/Ze (Cyrillic)/ü as options as described in section \ref{sec:Choice_ID_symbols} (it was used as a baseline in \ref{tab:selected_domains_rand_symb_options} and \ref{tab:selected_domain_rand_sym_content}). The deltas are calculated compared with  \ref{tab:selected_domains_baseline}. In this particular experiment, all models encountered a decline in accuracy, coupled with a significant increase in RStds values, except Llama-13b-chat. }
\label{tab:selected_domains_sym_set2}
\end{table} 

%% file: tables/selectedDomain_baseline_Style3.tex
\begin{table}[h!]
\centering
\resizebox{\linewidth}{!}{
\begin{tabular}{lcccc}
\hline
Model & Acc 0shot & RStd 0shot & Acc 5shot  & RStd 5shot \\ 
&($\Delta$Acc) &($\Delta$RStd)&($\Delta$Acc) &($\Delta$RStd) \\
\midrule
phi-2 & 28.3 (-9.3) & 6.0 (-0.7) & 34.6 (-6.3) & 5.7 (-1.04) \\
Yi-6B & 35.0 (-6.3) & 11.5 (1.4) & 39.0 (-1.7) & 13.5 (-0.6) \\
Mistral-7B & 34.3 (-4.7) & 10.7 (1.6) & 44.0 (3.0) & 16.3 (4.2) \\
Mistral-7B-Instruct & 35.0 (0.0) & 14.0 (0.7) & 38 (2.0) & 15.7 (0.0) \\
Llama-2-7b & 31.3 (2.0) & 12.6 (-1.0) & 32.6 (-0.7) & 16.9 (-0.7) \\
Llama-2-7b-chat & 27.0 (-5.3) & 12.5 (-7.3) & 32.6 (-0.7) & 13.3 (-8.0) \\
Llama-2-13b & 37.0 (0.7) & 14.0 (-3.0) & 40.0 (4.3) & 15.7 (1.9) \\
Llama-2-13b-chat & 33.0 (1.3) & 9.1 (-17.7) & 37.6 (5.0) & 17.33 (-7.4) \\
Yi-34B & 46.6 (-3.3) & 12.8 (1.4) & 47.6 (-1.7) & 10.1 (0.8) \\
Llama-2-70b & 41.3 (-0.7) & 10.5 (-4.0) & 49.0 (4.3) & 10.3 (4.2) \\
Llama-2-70b-chat & 39.3 (2.0) & 7.8 (-11.8) & 42.6 (1.7) & 11.9 (-6.5) \\

\midrule
$k_\tau$ & 0.564 && 0.6 & \\
\bottomrule
\end{tabular}
}

\caption{The average zero-shot results on the three selected domains baseline using the Hybrid style mentioned in section \ref{sec:Prompt_and_scoring_modifications}. The deltas are compared with \ref{tab:selected_domains_baseline} where the Rstd values exhibited a decrease and the accuracies remained relatively stable, except phi-2, which demonstrated the most significant decline in accuracy. }
\label{tab:selected_domains_baseline_style3}
\end{table}

%% file: tables/five_shot_option_fixing.tex
\begin{table}[h!]
\centering
\resizebox{\linewidth}{!}{
\begin{tabular}{lccccc}
\toprule
 Model & \textbf{Baseline} & \textbf{A} & \textbf{B} & \textbf{C} & \textbf{D} \\
\midrule
\texttt{phi-2} & 54.47 & \makecell{57.33 \\ \color{blue}{(+2.87)}} & \makecell{44.00 \\ \color{red}{(-10.47)}} & \makecell{25.00 \\ \color{red}{(-29.47)}} & \makecell{32.33 \\ \color{red}{(-22.13)}} \\
\texttt{Yi-6B} & 61.12 & \makecell{49.67 \\ \color{red}{(-11.45)}} & \makecell{23.67 \\ \color{red}{(-37.45)}} & \makecell{18.33 \\ \color{red}{(-42.78)}} & \makecell{44.67 \\ \color{red}{(-16.45)}} \\
\texttt{Mistral-7B} & 59.56 & \makecell{77.00 \\ \color{blue}{(+17.44)}} & \makecell{46.33 \\ \color{red}{(-13.23)}} & \makecell{48.33 \\ \color{red}{(-11.23)}} & \makecell{68.00 \\ \color{blue}{(+8.44)}} \\
\texttt{Mistral-7B-Instruct} & 53.48 & \makecell{78.33 \\ \color{blue}{(+24.85)}} & \makecell{42.33 \\ \color{red}{(-11.15)}} & \makecell{18.67 \\ \color{red}{(-34.82)}} & \makecell{49.33 \\ \color{red}{(-4.15)}} \\
\texttt{Llama-2-7b} & 41.81 & \makecell{79.00 \\ \color{blue}{(+37.19)}} & \makecell{57.33 \\ \color{blue}{(+15.52)}} & \makecell{24.67 \\ \color{red}{(-17.14)}} & \makecell{23.67 \\ \color{red}{(-18.14)}} \\
\texttt{Llama-2-7b-chat} & 46.37 & \makecell{16.67 \\ \color{red}{(-29.70)}} & \makecell{66.33 \\ \color{blue}{(+19.97)}} & \makecell{38.67 \\ \color{red}{(-7.70)}} & \makecell{14.33 \\ \color{red}{(-32.04)}} \\
\texttt{Llama-2-13b} & 52.08 & \makecell{33.67 \\ \color{red}{(-18.41)}} & \makecell{37.33 \\ \color{red}{(-14.75)}} & \makecell{45.33 \\ \color{red}{(-6.75)}} & \makecell{39.33 \\ \color{red}{(-12.75)}} \\
\texttt{Llama-2-13b-chat} & 53.12 & \makecell{20.00 \\ \color{red}{(-33.12)}} & \makecell{23.00 \\ \color{red}{(-30.12)}} & \makecell{61.33 \\ \color{blue}{(+8.21)}} & \makecell{15.67 \\ \color{red}{(-37.45)}} \\
\texttt{Yi-34B} & 73.38 & \makecell{59.00 \\ \color{red}{(-14.38)}} & \makecell{45.67 \\ \color{red}{(-27.71)}} & \makecell{53.67 \\ \color{red}{(-19.71)}} & \makecell{48.00 \\ \color{red}{(-25.38)}} \\
\bottomrule
\end{tabular}
}
\caption{Performance on five-shot MMLU when placing the correct answer at each possible position, for both the examples and the question asked. Similar to the zero-shot case mentioned in Section \ref{results}, all the LLMs tested showed a clear preference for specific positions/answer choice symbols, although the position varied among models and even in model families.}
\label{tab:five_shot_fixing}
\end{table}

%% file: tables/shuffle_full_options.tex
\begin{table}[h!]                                                                    
\centering                                                                           
\resizebox{\linewidth}{!}{                                                           
\begin{tabular}{lcccc}
\toprule
Model & Task acc & $\Delta$Acc & Task RStd & $\Delta$RStd \\
\midrule
\texttt{phi-2} & 51.01 & -3.45 & 8.82 & 4.82 \\
\texttt{Yi-6B} & 57.75 & -3.37 & 6.29 & 2.72 \\
\texttt{Mistral-7B} & 55.63 & -3.94 & 7.75 & 3.62 \\
\texttt{Mistral-7B-Instruct} & 52.09 & -1.39 & 4.02 & -0.57 \\
\texttt{Llama-2-7b} & 32.13 & -9.68 & 23.72 & 15.23 \\
\texttt{Llama-2-7b-chat} & 42.52 & -3.85 & 15.45 & -0.66 \\
\texttt{Llama-2-13b} & 48.24 & -3.84 & 8.29 & -3.75 \\
\texttt{Llama-2-13b-chat} & 51.83 & -1.29 & 5.24 & -7.56 \\
\texttt{Yi-34B} & 69.56 & -3.82 & 4.62 & -0.55 \\
\texttt{Llama-2-70b} & 63.32 & -2.12 & 3.33 & 0.13 \\
\texttt{Llama-2-70b-chat} & 58.80 & -2.31 & 1.91 & -9.04 \\
\bottomrule
\end{tabular}}

\caption{Reproducing shuffling ablation experiment from \cite{zheng2023large}. Randomly shuffling the order in which the options are presented. Surprisingly, all models demonstrated a decrease in accuracy, suggesting a lack of decisiveness in the experiment. However, these variations indicate a potential bias in the benchmark. }
\label{tab:shuffle_full_options}
\end{table}

%% file: tables/selectedDomains_randSymContent.tex
\begin{table}[h!]
    \centering
    \resizebox{\linewidth}{!}{%
            \begin{tabular}{lcc}
                \toprule
                Model & Task Acc ($\Delta$Acc)& Task RStd ($\Delta$RStd) \\
                \midrule
                \texttt{phi-2} & 25.33 (-12.33)	& 42.35 (35.57) \\
                \texttt{Yi-6B} & 30.66 (-2.0) & 26.68 (3.8) \\
                \texttt{Mistral-7B} & 34.00 (-1.3) & 22.37 (3.6) \\
                \texttt{Mistral-7B-Instruct} & 30.33 (-3.7) & 23.08 (-3.9) \\
                \texttt{Llama-2-7b} & 25.66 (-4.0) & 24.98 (-0.9) \\
                \texttt{Llama-2-7b-chat} & 28.00 (-3.3) & 28.49 (1.4) \\
                \texttt{Llama-2-13b} & 29.33 (-5.0) & 26.75 (0.6) \\
                \texttt{Llama-2-13b-chat} & 29.66 (-4.3) & 20.58 (-1.4) \\
                \texttt{Yi-34B} & 36.00 (-6.7) & 19.48 (-3.3) \\
                \texttt{Llama-2-70b} & 37.33 (-2.3) & 10.41 (-4.7) \\
                \texttt{Llama-2-70b-chat} & 31.66 (-4.3) & 23.25 (-6.3) \\
                \midrule
$k_\tau$ = 0.564 \\
                \bottomrule
            \end{tabular}
        }
    
\caption{The average zero-shot results on the three selected domains using Symbols Set2 (mentioned in section \ref{sec:Choice_ID_symbols} and shuffling the choices while fixing the order of the choices symbols. The deltas are measured compared with  \ref{tab:selected_domains_sym_set2}. As displayed in the table, mostly all the models faced a decrease in accuracy while the RStds values were not decisive. The most affected model in this experiment was phi-2.}
\label{tab:selected_domain_rand_sym_content}
\end{table}

%% file: tables/selectedDomains_randSymOptions.tex
\begin{table}[h!]
    \centering
    \resizebox{\linewidth}{!}{%
            \begin{tabular}{lcc}
                \toprule
                Model & Task Avg Acc ($\Delta$Acc)& Task Avg RStd ($\Delta$RStd) \\
                \midrule
                \texttt{phi-2} & 29.00(-8.6)	& 12.4 (5.6) \\
                \texttt{Yi-6B} & 34.67 (2.0) & 22.84 (0.0) \\
                \texttt{Mistral-7B} & 29.33 (-6.0) & 16.52 (-2.3) \\
                \texttt{Mistral-7B-Instruct} & 28.33 (-5.7) & 22.10 (-4.9) \\
                \texttt{Llama-2-7b} & 26.67 (-3.0) & 28.62 (2.8) \\
                \texttt{Llama-2-7b-chat} & 32.00 (0.7) & 15.64 (-11.4) \\
                \texttt{Llama-2-13b} & 26.33 (-8.0) & 21.31 (-4.8) \\
                \texttt{Llama-2-13b-chat} & 34.00 (0.0) & 16.97 (-5.0) \\
                \texttt{Yi-34B} & 41.00 (-1.7) & 20.28 (-2.5) \\
                \texttt{Llama-2-70b} & 38.67 (-1.0) & 7.65 (-7.5) \\
                \texttt{Llama-2-70b-chat} & 40.33 (4.3) & 15.78 (-13.8) \\
                 \midrule
$k_\tau$ = 0.455 \\
                \bottomrule
            \end{tabular}
        }

\caption{The average zero-shot results on the three selected domains using Symbols Set2 mentioned in section \ref{sec:Choice_ID_symbols}. This experiment focused on shuffling the symbols while maintaining the original listing order of the choices. Compared with \ref{tab:selected_domains_sym_set2},Most of the models were impacted in terms of accuracy and RStds, indicating that randomization affects the models even after changing the symbols. }
\label{tab:selected_domains_rand_symb_options}
\end{table}

%% file: tables/selectedDomains_rand_style3.tex
\begin{table}[h!]
    \centering
    \resizebox{\linewidth}{!}{%
            \begin{tabular}{lcccc}
                \toprule
Model & Acc 0shot & RStd 0shot  & Acc 5shot  & RStd 5shot  \\
&($\Delta$Acc) & ($\Delta$RStd) & ($\Delta$Acc) &($\Delta$RStd)\\

\midrule
\texttt{phi-2} & 30.6 (2.3) &12.8 (6.8) & 32.6(-2) &13.6 (7.8) \\
\texttt{Yi-6B} & 30.3 (-4.7) & 12.0 (0.5) & 34.3 (-4.7) & 11.4 (-2.1) \\
\texttt{Mistral-7B} & 31.6 (-2.7) & 12.5 (1.8) & 39 (-5) & 11.1 (-5.2) \\
\texttt{Mistral-7B-Instruct} & 32.66 (-2.3) & 11.18 (-2.9) & 37 (-1) & 7.94 (-7.8) \\
\texttt{Llama-2-7b} & 28.6 (-2.7) & 11.4 (-1.2) & 33.3 (0.7) & 15.1 (-1.8) \\
\texttt{Llama-2-7b-chat} & 29.3 (2.3) & 16.2 (3.7) & 35 (2.3) & 16.6 (3.3) \\
\texttt{Llama-2-13b} & 35.3 (-1.7) & 10.1 (-3.9) & 37.6 (-2.3) & 12.1 (-3.6) \\
\texttt{Llama-2-13b-chat} & 29.6 (-3.3) & 10.9 (1.9) & 35.3 (-2.3) & 17.0 (-0.3) \\
\texttt{Yi-34B} & 43 (-3.7) & 5.4 (-7.4) & 48.3 (0.7) & 11.7 (1.6) \\
\texttt{Llama-2-70b} & 40 (-1.3) & 9.0 (-1.5) & 48 (-1) & 10.5 (0.1) \\
\texttt{Llama-2-70b-chat} & 35 (-4.3) & 11.1 (3.4) & 41.3 (-1.3) & 6.8 (-5.1) \\

                \midrule
$k_\tau$ &  0.527 &&0.382& \\
                \bottomrule
            \end{tabular}
        }

\caption{The selected domains results after randomizing the choices using Hybrid style mentioned in section \ref{sec:Prompt_and_scoring_modifications}, the deltas are calculated from this table \ref{tab:selected_domains_baseline_style3} where it showed more consistency compared to the results of other randomization settings (\ref{tab:selected_domains_rand},\ref{tab:selected_domain_rand_sym_content}, and \ref{tab:selected_domains_rand_symb_options}).}
\label{tab:selected_domains_rand_style3}
\end{table}

%% file: tables/mmlu_symbols_set1_options_0shot.tex
\begin{table}[h!]
\centering
\resizebox{\linewidth}{!}{
\begin{tabular}{lcccc}
\toprule
                       Model &  Task acc &  $\Delta$Acc &  Task RStd &  $\Delta$RStd \\
\midrule
\texttt{phi-2} & 31.92 & -22.55 & 20.23 & 16.22 \\
              \texttt{Yi-6B} &     46.87 &       -14.25 &      15.24 &         11.67 \\
         \texttt{Mistral-7B} &     42.68 &       -16.88 &      29.07 &         24.94 \\
\texttt{Mistral-7B-Instruct} &     47.90 &        -5.58 &      15.06 &         10.48 \\
         \texttt{Llama-2-7b} &     26.23 &       -15.58 &      33.78 &         25.29 \\
    \texttt{Llama-2-7b-chat} &     41.01 &        -5.36 &      14.17 &         -1.94 \\
        \texttt{Llama-2-13b} &     41.05 &       -11.03 &      23.54 &         11.50 \\
   \texttt{Llama-2-13b-chat} &     48.09 &        -5.03 &      20.82 &          8.02 \\
             \texttt{Yi-34B} &     66.56 &        -6.82 &      10.13 &          4.96 \\
        \texttt{Llama-2-70b} &     57.94 &        -7.50 &      16.52 &         13.32 \\
   \texttt{Llama-2-70b-chat} &     59.00 &        -2.11 &      10.09 &         -0.86 \\
   %\texttt{allam-llama-7b-sft} & 48.017 & -5.563  & &  \\
      \midrule
$k_\tau$ = 0.6 \\
\bottomrule
\end{tabular}

}

\caption{The zero-shot results of MMLU on Symbols Set1 mentioned in section \ref{sec:Choice_ID_symbols}. All of the models demonstrated reduced accuracies, while most of them showed an increase in RStds values compared with \ref{tab:mmlu_baselines}.}
\label{tab:mmlu_symbols_set1_options_0shot}
\end{table}

%% file: tables/mmlu_symbols_set2_options_0shot.tex
\begin{table}[h]

\resizebox{\linewidth}{!}{
\begin{tabular}{lcccc}
\toprule
                       Model &  Task acc &  $\Delta$Acc &  Task RStd &  $\Delta$RStd \\
\midrule
\texttt{phi-2} & 29.85 & -24.62 & 39.10 & 35.09 \\
              \texttt{Yi-6B} &     47.58 &       -13.54 &      26.09 &         22.52 \\
         \texttt{Mistral-7B} &     52.63 &        -6.94 &      15.87 &         11.74 \\
\texttt{Mistral-7B-Instruct} &     48.33 &        -5.15 &      18.70 &         14.12 \\
         \texttt{Llama-2-7b} &     29.76 &       -12.05 &      32.09 &         23.60 \\
    \texttt{Llama-2-7b-chat} &     43.34 &        -3.03 &      18.20 &          2.09 \\
        \texttt{Llama-2-13b} &     42.06 &       -10.02 &      23.75 &         11.70 \\
   \texttt{Llama-2-13b-chat} &     49.23 &        -3.89 &      14.07 &          1.28 \\
             \texttt{Yi-34B} &     67.03 &        -6.35 &      12.48 &          7.31 \\
        \texttt{Llama-2-70b} &     62.60 &        -2.84 &       3.21 &          0.01 \\
   \texttt{Llama-2-70b-chat} &     57.01 &        -4.10 &      18.53 &          7.59 \\
   \midrule
$k_\tau$ = 0.636 \\
\bottomrule
\end{tabular}

}
\caption{ The zero-shot results of MMLU on Symbols Set2 mentioned in section \ref{sec:Choice_ID_symbols}. Compared with the original MMLU implementation that used A/B/C/D as symbols(\ref{tab:mmlu_baselines}), the majority of models in this experiment had notably lower accuracies while the RStd values increased.}
\label{tab:mmlu_symbols_set2_options_0shot}
\end{table}

%% file: tables/mmlu_style2_0shot.tex
\begin{table}[h!]
\centering
\resizebox{\linewidth}{!}{
\begin{tabular}{lcccc}
\toprule
                       Model &  Task acc &  $\Delta$Acc &  Task RStd &  $\Delta$RStd \\
\midrule
              \texttt{phi-2} & 40.714 & -13.751 & 1.398 & -2.607 \\
              \texttt{Yi-6B} &     42.40 &       -18.72 &       1.49 &         -2.08 \\
         \texttt{Mistral-7B} &     45.69 &       -13.87 &       1.26 &         -2.87 \\
\texttt{Mistral-7B-Instruct} &     43.51 &        -9.98 &       1.53 &         -3.05 \\
         \texttt{Llama-2-7b} &     40.81 &        -1.00 &       1.19 &         -7.30 \\
    \texttt{Llama-2-7b-chat} &     40.44 &        -5.93 &       1.79 &        -14.32 \\
        \texttt{Llama-2-13b} &     44.09 &        -7.99 &       1.29 &        -10.75 \\
   \texttt{Llama-2-13b-chat} &     43.87 &        -9.25 &       1.86 &        -10.93 \\
             \texttt{Yi-34B} &     49.33 &       -24.05 &       3.76 &         -1.41 \\
        \texttt{Llama-2-70b} &     48.74 &       -16.70 &       0.99 &         -2.21 \\
   \texttt{Llama-2-70b-chat} &     46.34 &       -14.77 &       1.57 &         -9.38 \\
   \midrule
$k_\tau$ = 0.527\\
\bottomrule
\end{tabular}

}
\caption{The zero-shot results of MMLU using the Cloze style mentioned in \ref{sec:Prompt_and_scoring_modifications}. As anticipated, employing this style led to significantly low RStd values compared with the Symbols scoring style in table \ref{tab:mmlu_baselines}, but it also had a considerable impact on accuracy, resulting in a noticeable decrease in most models.}
\label{tab:mmlu_style2_choices_0shot}
\end{table}

%% file: tables/mmlu_style3_0shot.tex
\begin{table}[h!]
\centering
\resizebox{\linewidth}{!}{
\begin{tabular}{lcccc}
\toprule
                       Model &  Task acc &  $\Delta$Acc &  Task RStd &  $\Delta$RStd \\
\midrule
              \texttt{phi-2} &     38.47 &       -16.01 &       2.55 &         -1.45 \\
              \texttt{Yi-6B} &     44.90 &       -16.22 &       3.80 &          0.23 \\
         \texttt{Mistral-7B} &     42.94 &       -16.62 &       5.09 &          0.96 \\
\texttt{Mistral-7B-Instruct} &     39.27 &       -14.21 &       3.46 &         -1.12 \\
         \texttt{Llama-2-7b} &     37.79 &        -4.02 &       3.79 &         -4.70 \\
    \texttt{Llama-2-7b-chat} &     37.68 &        -8.69 &       3.52 &        -12.58 \\
        \texttt{Llama-2-13b} &     43.88 &        -8.20 &       5.14 &         -6.91 \\
   \texttt{Llama-2-13b-chat} &     39.14 &       -13.98 &       4.82 &         -7.98 \\
             \texttt{Yi-34B} &     59.52 &       -13.86 &       2.63 &         -2.54 \\
        \texttt{Llama-2-70b} &     55.11 &       -10.33 &       2.00 &         -1.20 \\
   \texttt{Llama-2-70b-chat} &     47.26 &       -13.85 &       3.35 &         -7.60 \\
         \midrule
$k_\tau$ = 0.709 \\
\bottomrule
\end{tabular}

}
\caption{The zero-shot results of MMLU using the Hybrid style mentioned in \ref{sec:Prompt_and_scoring_modifications}. This style resulted in decreased accuracy but demonstrated more stability and lower RStd values when comparing it with the Symbols scoring style \ref{tab:mmlu_baselines}. This style may help reduce the selection and token bias seen in prior experiments.}
\label{tab:mmlu_style3}
\end{table}

%% file: tables/arc_c_0shot_Style1.tex
\begin{table}[h]
    \centering
    \resizebox{\linewidth}{!}{%
            \begin{tabular}{lcc}
                \toprule
                Model & Task Acc ($\Delta$Acc) & Task RStd ($\Delta$RStd)\\
                \midrule
                \texttt{phi-2} & 76.8 (22.7) & 4.2 (1.6) \\
                \texttt{Yi-6B} & 78.3 (27.8) & 2.6 (0.5) \\
                \texttt{Mistral-7B} & 74.8 (21.2) & 5.9 (3.3) \\
                \texttt{Mistral-7B-Instruct} & 69.3 (17.3) & 4.3 (2.9) \\
                \texttt{Llama-2-7b} & 42.4 (-3.9) & 14.1 (10.0) \\
                \texttt{Llama-2-7b-chat} & 57.6 (13.3) & 13.8 (11.6) \\
                \texttt{Llama-2-13b} & 62.0 (13.0) & 8.9 (6.0) \\
                \texttt{Llama-2-13b-chat} & 65.3 (15.1) & 12.7 (10.9) \\
                \texttt{Yi-34B} & 90.7 (29.1) & 0.5 (-1.9) \\
                \texttt{Llama-2-70b} & 81.9 (24.7) & 2.6 (0.025) \\
                \texttt{Llama-2-70b-chat} & 78.4 (24.1) & 6.8 (5.3) \\
                \midrule
$k_\tau$ = 0.855 \\
                \bottomrule
                
            \end{tabular}
        }

\caption{The table displays the results of zero-shot on ARC-C with Symbols scoring style mentioned in \ref{sec:Prompt_and_scoring_modifications}. Compared with \ref{tab:arc_baselines}, all models, except Llama-2-7b, showed higher accuracies. An increase in Rstds values was observed, particularly in the Llama-2 7b, 7b-chat, and 13b models. This proves that if we provide choices in the prompt, models will perform better.}

\label{tab:arc_c_mmlu_0shot}
\end{table}

%% file: tables/arc_c_0shot_Style3.tex
\begin{table}[h]
    \centering
    \resizebox{\linewidth}{!}{%
            \begin{tabular}{lcc}
                \toprule
                Model & Task Acc ($\Delta$Acc) & Task RStd ($\Delta$RStd)\\
                \midrule
                \texttt{phi-2} & 58.4 (4.3) & 4.9 (2.4) \\
                \texttt{Yi-6B} & 59.9 (9.4) & 8.4 (6.3) \\
                \texttt{Mistral-7B} & 52.6 (-0.9) & 6.9 (4.3) \\
                \texttt{Mistral-7B-Instruct} & 54.1 (2.1) & 4.3 (2.9) \\
                \texttt{Llama-2-7b} & 38.7 (-7.5) & 7.8 (3.7) \\
                \texttt{Llama-2-7b-chat} & 46.6 (2.3) & 2.7 (0.5) \\
                \texttt{Llama-2-13b} & 52.4 (3.4) & 9.1 (6.1) \\
                \texttt{Llama-2-13b-chat} & 53.5 (3.3) & 4.6 (2.7) \\
                \texttt{Yi-34B} & 83.0 (21.5) & 3.8 (1.2) \\
                \texttt{Llama-2-70b} & 72.6 (15.4) & 3.9 (1.3) \\
                \texttt{Llama-2-70b-chat} & 64.7 (10.4) & 4.2 (2.7) \\
                                \midrule
$k_\tau$ = 0.782 \\
                \bottomrule
            \end{tabular}
        }
    
    \caption{The zero-shot results of ARC-C using the Hybrid style discussed in \ref{sec:Prompt_and_scoring_modifications}. In some models, it exhibits higher accuracy than the baseline (Table \ref{tab:arc_baselines}) and more stable RStd values (compared to \ref{tab:arc_c_mmlu_0shot}). The deltas are calculated using this table \ref{tab:arc_baselines}.}
    \label{tab:label}
\end{table}

%% file: tables/trivial_examples_exp1.tex
% \begin{table}[ht]
%     \centering
%     \resizebox{\linewidth}{!}{
%     {\color{violet}\begin{tabular}{ l c c c c c }
%     \toprule
%      \multirow{2}{*}{Model} & \multicolumn{2}{c}{Baseline} & \multicolumn{3}{c}{Trivial Examples} \\
%     \cmidrule(lr){2-3} \cmidrule(lr){4-6}
%            & 0-shot & 5-shot & V1 & V2 & V3 \\
%     \midrule
%     \texttt{Yi\_6b\_base}      & 62.1 & 63.8 & 61.2 & 61.4 & 61.5 \\
%     \texttt{Llama2\_7b\_base}  & 42.7 & 46.6 & 43.1 & 44.1 & 44.9 \\
%     \texttt{Llama2\_7b\_chat}  & 47.0 & 48.2 & 47.1 & 47.8 & 48.0 \\
%     \texttt{Mistral\_7b\_base} & 61.4 & 63.5 & 59.8 & 60.5 & 60.5 \\
%     \texttt{Mistral\_7b\_chat} & 54.9 & 55.5 & 52.5 & 52.0 & 51.7 \\
%     \texttt{Llama2\_13b\_base} & 52.9 & 55.3 & 52.5 & 53.0 & 53.0 \\
%     \texttt{Llama2\_13b\_chat} & 53.8 & 54.6 & 51.7 & 52.9 & 52.9 \\
%     \texttt{Yi\_34b\_base}     & 74.7 & 76.3 & 73.4 & 73.7 & 74.0 \\
%     \texttt{Llama2\_70b\_base} & 66.2 & 69.6 & 66.3 & 66.3 & 66.6 \\
%     \texttt{Llama2\_70b\_chat} & 62.3 & 63.9 & 60.9 & 61.4 & 61.1 \\
%     \bottomrule
%     \end{tabular}}}
%     \caption{\color{violet}Trivial examples as a few-shot performance comparison. Baseline results (0-shot and 5-shot) along with the few-shot results (Version 1, 2, and 3).}
%     \label{tab:trivial_examples}
% \end{table}

\begin{table}[h!]
\centering
\resizebox{\linewidth}{!}{
\begin{tabular}{lcccc}
\toprule
                            Model &  Task acc &  $\Delta$Acc \\
\midrule
                   \texttt{phi-2} &     54.21 &        -0.26 \\
                   \texttt{Yi-6B} &     60.11 &        -1.00 \\
              \texttt{Mistral-7B} &     58.45 &        -1.11 \\
     \texttt{Mistral-7B-Instruct} &     51.14 &        -2.34 \\
              \texttt{Llama-2-7b} &     42.77 &         0.96 \\
         \texttt{Llama-2-7b-chat} &     46.35 &        -0.02 \\
             \texttt{Llama-2-13b} &     51.72 &        -0.36 \\
        \texttt{Llama-2-13b-chat} &     50.94 &        -2.18 \\
                  \texttt{Yi-34B} &     72.28 &        -1.10 \\
             \texttt{Llama-2-70b} &     65.25 &        -0.18 \\
        \texttt{Llama-2-70b-chat} &     59.79 &        -1.32 \\
\midrule
$k_\tau$ = 0.927 \\
\bottomrule
\end{tabular}

}
\caption{Trivial examples few-shot results using the version 1 examples with respect to zero-shot baseline accuracy.}
\label{tab:trivial_examples_exp1}
\end{table}

%% file: tables/trivial_examples_exp2.tex
\begin{table}[h]
\centering
\resizebox{\linewidth}{!}{
\begin{tabular}{lcccc}
\toprule
                            Model &  Task acc &  $\Delta$Acc \\
\midrule
                   \texttt{phi-2} &     53.18 &        -1.28 \\
                   \texttt{Yi-6B} &     60.28 &        -0.84 \\
              \texttt{Mistral-7B} &     59.41 &        -0.15 \\
     \texttt{Mistral-7B-Instruct} &     50.95 &        -2.53 \\
              \texttt{Llama-2-7b} &     43.52 &         1.71 \\
         \texttt{Llama-2-7b-chat} &     46.82 &         0.46 \\
             \texttt{Llama-2-13b} &     52.51 &         0.44 \\
        \texttt{Llama-2-13b-chat} &     51.84 &        -1.27 \\
                  \texttt{Yi-34B} &     72.29 &        -1.09 \\
             \texttt{Llama-2-70b} &     65.13 &        -0.31 \\
        \texttt{Llama-2-70b-chat} &     60.28 &        -0.83 \\
\midrule
$k_\tau$ = 0.891 \\
\bottomrule
\end{tabular}

}
\caption{Trivial examples few-shot results with version 2 examples with respect to zero-shot baseline accuracy.}
\label{tab:trivial_examples_exp2}
\end{table}

%% file: tables/trivial_examples_exp3.tex
\begin{table}[h]
\centering
\resizebox{\linewidth}{!}{
\begin{tabular}{lccc}
\toprule
                            Model &  Task acc &  $\Delta$Acc \\
\midrule
                   \texttt{phi-2} &     53.22 &        -1.25 \\
                   \texttt{Yi-6B} &     60.46 &        -0.66 \\
              \texttt{Mistral-7B} &     59.27 &        -0.30 \\
     \texttt{Mistral-7B-Instruct} &     50.58 &        -2.91 \\
              \texttt{Llama-2-7b} &     44.47 &         2.66 \\
         \texttt{Llama-2-7b-chat} &     46.90 &         0.53 \\
             \texttt{Llama-2-13b} &     52.24 &         0.16 \\
        \texttt{Llama-2-13b-chat} &     51.88 &        -1.24 \\
                  \texttt{Yi-34B} &     73.16 &        -0.22 \\
             \texttt{Llama-2-70b} &     65.42 &        -0.02 \\
        \texttt{Llama-2-70b-chat} &     60.02 &        -1.09 \\
\midrule
$k_\tau$ = 0.891 \\
\bottomrule
\end{tabular}

}
\caption{Trivial examples few-shot results with version 3 examples, with respect to zero-shot baseline accuracy.}
\label{tab:trivial_examples_exp3} 
\end{table}

%% file: tables/varying_instructions_prompts.tex
\begin{table}[h]
\centering
\resizebox{\linewidth}{!}{
\begin{tabular}{lcccc}
\toprule
                            Model &  Task acc &  $\Delta$Acc &  Task RStd &  $\Delta$RStd \\
\midrule
                   \texttt{phi-2} &     53.92 &        -0.54 &       4.07 &          0.07 \\
                   \texttt{Yi-6B} &     60.80 &        -0.31 &       3.43 &         -0.14 \\
              \texttt{Mistral-7B} &     59.02 &        -0.54 &       3.73 &         -0.40 \\
     \texttt{Mistral-7B-Instruct} &     53.29 &        -0.19 &       4.74 &          0.16 \\
              \texttt{Llama-2-7b} &     41.80 &        -0.01 &       4.51 &         -3.99 \\
         \texttt{Llama-2-7b-chat} &     46.68 &         0.31 &      14.93 &         -1.17 \\
             \texttt{Llama-2-13b} &     51.92 &        -0.16 &      12.05 &          0.00 \\
        \texttt{Llama-2-13b-chat} &     53.27 &         0.15 &      12.83 &          0.03 \\
                  \texttt{Yi-34B} &     72.94 &        -0.44 &       5.52 &          0.35 \\
             \texttt{Llama-2-70b} &     64.83 &        -0.60 &       2.81 &         -0.40 \\
        \texttt{Llama-2-70b-chat} &     61.14 &         0.03 &      10.94 &         -0.00 \\
\midrule
$k_\tau$=0.964 \\
\bottomrule
\end{tabular}
}
\caption{Zero-shot results of removing the subject name from the prompt. (experiment 1 from figure 
 \ref{fig:varying_inst_prompts_shaykha}). There are minimal changes in performance when applying this perturbation.}
\label{tab:label_prompt_modification1}
\end{table}

% \begin{table}[ht]
% \centering
% \resizebox{\linewidth}{!}{
% \begin{tabular}{lcccc}
% \toprule
%                             Model &  Task acc &  $\Delta$Acc &  Task RStd &  $\Delta$RStd \\
% \midrule
%                    \texttt{phi-2} &     54.70 &         0.23 &       4.06 &          0.06 \\
%                    \texttt{Yi-6B} &     61.24 &         0.13 &       1.74 &         -1.83 \\
%          \texttt{Mistral-7B-v0.1} &     59.71 &         0.15 &       8.32 &          4.19 \\
% \texttt{Mistral-7B-Instruct-v0.1} &     53.40 &        -0.08 &       6.26 &          1.68 \\
%            \texttt{Llama-2-7b-hf} &     43.26 &         1.45 &       9.63 &          1.13 \\
%       \texttt{Llama-2-7b-chat-hf} &     45.95 &        -0.42 &      15.42 &         -0.68 \\
%           \texttt{Llama-2-13b-hf} &     51.77 &        -0.31 &      15.68 &          3.64 \\
%      \texttt{Llama-2-13b-chat-hf} &     52.60 &        -0.52 &      14.04 &          1.24 \\
%                   \texttt{Yi-34B} &     73.40 &         0.02 &       5.06 &         -0.11 \\
%           \texttt{Llama-2-70b-hf} &     65.22 &        -0.22 &       6.05 &          2.85 \\
%      \texttt{Llama-2-70b-chat-hf} &     60.68 &        -0.43 &      11.84 &          0.89 \\
% \midrule
% $k_\tau$=1.0 \\
% \bottomrule
% \end{tabular}
% }
% \caption{Zero-shot results of experiment 2 (zero-shot specific format)}
% \label{tab:label_prompt_modification2}
% \end{table}

\begin{table}[ht]
\centering
\resizebox{\linewidth}{!}{
\begin{tabular}{lcccc}
\toprule
                            Model &  Task acc &  $\Delta$Acc &  Task RStd &  $\Delta$RStd \\
\midrule
                   \texttt{phi-2} &     54.21 &        -0.26 &       4.21 &          0.20 \\
                   \texttt{Yi-6B} &     61.06 &        -0.06 &       2.33 &         -1.24 \\
              \texttt{Mistral-7B} &     60.16 &         0.60 &       2.08 &         -2.06 \\
     \texttt{Mistral-7B-Instruct} &     53.67 &         0.19 &       4.03 &         -0.56 \\
              \texttt{Llama-2-7b} &     41.42 &        -0.39 &      15.05 &          6.56 \\
         \texttt{Llama-2-7b-chat} &     47.22 &         0.85 &      14.22 &         -1.88 \\
             \texttt{Llama-2-13b} &     53.46 &         1.38 &      10.46 &         -1.59 \\
        \texttt{Llama-2-13b-chat} &     53.20 &         0.08 &      11.09 &         -1.71 \\
                  \texttt{Yi-34B} &     73.64 &         0.26 &       5.68 &          0.51 \\
             \texttt{Llama-2-70b} &     65.48 &         0.04 &       3.51 &          0.30 \\
        \texttt{Llama-2-70b-chat} &     61.20 &         0.09 &      10.31 &         -0.63 \\
\midrule
$k_\tau$=0.927 \\   
\bottomrule
\end{tabular}
}
\caption{Zero-shot results on adding the ``Correct" token in the prompt. (experiment 2 from figure 
 \ref{fig:varying_inst_prompts_shaykha}). There are minimal changes in performance when applying this perturbation.}
\label{tab:label_prompt_modification2}
\end{table}

%%%%%%%%%%%%%%%%%%%%%%%%%%%%%%%%%%%%%%%%%%%%%%%%%%%%%%%%%%%%%%%%%%%%%%%%%%%%%%%%%%%%%%%%%
\begin{table}[ht]
\centering
\resizebox{\linewidth}{!}{
\begin{tabular}{lcccc}
\toprule
                            Model &  Task acc &  $\Delta$Acc &  Task RStd &  $\Delta$RStd \\
\midrule
                   \texttt{phi-2} &     56.69 &        -0.08 &       2.57 &         -0.08 \\
                   \texttt{Yi-6B} &     63.69 &         0.46 &       3.22 &          0.68 \\
              \texttt{Mistral-7B} &     62.60 &         0.23 &       2.98 &          1.33 \\
     \texttt{Mistral-7B-Instruct} &     53.99 &         0.04 &       4.62 &         -0.16 \\
              \texttt{Llama-2-7b} &     45.80 &        -0.09 &       8.75 &         -0.17 \\
         \texttt{Llama-2-7b-chat} &     47.42 &         0.20 &      12.03 &         -0.11 \\
             \texttt{Llama-2-13b} &     55.47 &         0.41 &       5.04 &          0.62 \\
        \texttt{Llama-2-13b-chat} &     53.58 &         0.05 &       8.32 &          0.00 \\
                  \texttt{Yi-34B} &     76.36 &        -0.02 &       2.14 &         -0.02 \\
             \texttt{Llama-2-70b} &     68.71 &        -0.07 &       1.63 &          0.06 \\
        \texttt{Llama-2-70b-chat} &     63.14 &        -0.03 &       8.49 &          0.43 \\
\midrule
$k_\tau$=1.0 \\
\bottomrule
\end{tabular}
}
\caption{Few-shot results of removing the subject name from the prompt. (experiment 1 from figure 
 \ref{fig:varying_inst_prompts_shaykha}). There are minimal changes in performance when applying this perturbation.}
\label{tab:label_prompt_modification3}
\end{table}

% \begin{table}[ht]
% \centering
% \resizebox{\linewidth}{!}{
% {\color{purple}
% \begin{tabular}{lcccc}
% \toprule
%                             Model &  Task acc &  $\Delta$Acc &  Task RStd &  $\Delta$RStd \\
% \midrule
%                    \texttt{phi-2} &     56.77 &         0.00 &       2.65 &          0.00 \\
%                    \texttt{Yi-6B} &     63.55 &         0.33 &       3.13 &          0.59 \\
%          \texttt{Mistral-7B-v0.1} &     62.64 &         0.28 &       2.95 &          1.31 \\
% \texttt{Mistral-7B-Instruct-v0.1} &     54.05 &         0.10 &       4.41 &         -0.37 \\
%            \texttt{Llama-2-7b-hf} &     45.83 &        -0.05 &       8.75 &         -0.17 \\
%       \texttt{Llama-2-7b-chat-hf} &     47.22 &         0.00 &      12.17 &          0.03 \\
%           \texttt{Llama-2-13b-hf} &     55.21 &         0.15 &       4.73 &          0.30 \\
%      \texttt{Llama-2-13b-chat-hf} &     53.60 &         0.08 &       8.33 &          0.02 \\
%                   \texttt{Yi-34B} &     76.44 &         0.06 &       2.43 &          0.27 \\
%           \texttt{Llama-2-70b-hf} &     68.78 &         0.00 &       1.56 &          0.00 \\
%      \texttt{Llama-2-70b-chat-hf} &     63.17 &         0.00 &       8.06 &          0.00 \\
% \midrule
% $k_\tau$=1.0 \\
% \bottomrule
% \end{tabular}

% }
% }
% \caption{\color{purple}Few-shot results of experiment 2 (zero-shot specific format)}
% \label{tab:label}
% \end{table}

\begin{table}[ht]
\centering
\resizebox{\linewidth}{!}{

\begin{tabular}{lcccc}
\toprule
                            Model &  Task acc &  $\Delta$Acc &  Task RStd &  $\Delta$RStd \\
\midrule
                   \texttt{phi-2} &     56.57 &        -0.21 &       3.95 &          1.30 \\
                   \texttt{Yi-6B} &     63.20 &        -0.03 &       4.01 &          1.47 \\
              \texttt{Mistral-7B} &     62.79 &         0.43 &       3.51 &          1.87 \\
     \texttt{Mistral-7B-Instruct} &     53.85 &        -0.10 &       5.51 &          0.73 \\
              \texttt{Llama-2-7b} &     46.21 &         0.33 &       7.14 &         -1.78 \\
         \texttt{Llama-2-7b-chat} &     47.48 &         0.26 &      10.42 &         -1.73 \\
             \texttt{Llama-2-13b} &     55.18 &         0.11 &       4.79 &          0.37 \\
        \texttt{Llama-2-13b-chat} &     53.75 &         0.23 &       6.58 &         -1.74 \\
                  \texttt{Yi-34B} &     75.98 &        -0.41 &       1.71 &         -0.46 \\
             \texttt{Llama-2-70b} &     69.10 &         0.32 &       0.83 &         -0.73 \\
        \texttt{Llama-2-70b-chat} &     62.86 &        -0.31 &       7.20 &         -0.86 \\
\midrule
$k_\tau$=1.0 \\
\bottomrule
\end{tabular}
}
\caption{Few-shot results on adding the ``Correct" token in the prompt. (experiment 2 from figure 
 \ref{fig:varying_inst_prompts_shaykha}). There are minimal changes in performance when applying this perturbation.}
\label{tab:label_prompt_modification5}
\end{table}

%% file: tables/fewshot_subject_indep.tex
\begin{table}[h!]
\centering
\resizebox{\linewidth}{!}{
\begin{tabular}{lcc}
\toprule
                            Model &  Task acc &  $\Delta$Acc\\% &  Task RStd &  $\Delta$RStd \\
\midrule
                   \texttt{phi-2} &     54.94 &        -1.84\\% &       1.93 &         -0.72 \\
                   \texttt{Yi-6B} &     61.51 &        -1.72\\% &       3.25 &          0.71 \\
              \texttt{Mistral-7B} &     59.56 &        -2.80\\% &       2.98 &          1.33 \\
     \texttt{Mistral-7B-Instruct} &     51.72 &        -2.24\\% &       4.88 &          0.10 \\
              \texttt{Llama-2-7b} &     44.10 &        -1.79\\% &      10.70 &          1.78 \\
         \texttt{Llama-2-7b-chat} &     46.92 &        -0.30\\% &      13.71 &          1.56 \\
             \texttt{Llama-2-13b} &     52.61 &        -2.46\\% &       8.17 &          3.75 \\
        \texttt{Llama-2-13b-chat} &     52.63 &        -0.90\\% &      11.47 &          3.15 \\
                  \texttt{Yi-34B} &     73.89 &        -2.50\\% &       4.55 &          2.38 \\
             \texttt{Llama-2-70b} &     66.26 &        -2.52\\% &       3.83 &          2.26 \\
        \texttt{Llama-2-70b-chat} &     60.85 &        -2.31\\% &      10.64 &          2.58 \\
\midrule
$k_\tau$ = 0.927 &&\\
\bottomrule
\end{tabular}
}
\caption{Subject independent five-shots example results with the first prompt. (follow Figure \ref{fig:subject_independent_prompts} for details). With few exceptions, most models exhibit a 2\% drop from changing the few shots example domains. For models that are not fine-tuned, we noticed a performance that is halfway between the standard zero-shot and five-shot. Indicating that these models utilize the few shots for both formatting and knowledge domain information.}
\label{tab:subject_indep_fewshot1}
\end{table}

\begin{table}[h!]
\centering
\resizebox{\linewidth}{!}{
\begin{tabular}{lcc}
\toprule
                            Model &  Task acc &  $\Delta$Acc\\% &  Task RStd &  $\Delta$RStd \\
\midrule
                   \texttt{phi-2} &     55.25 &        -1.52\\% &       1.97 &         -0.68 \\
                   \texttt{Yi-6B} &     61.15 &        -2.08\\% &       3.43 &          0.89 \\
              \texttt{Mistral-7B} &     59.68 &        -2.69\\% &       5.40 &          3.75 \\
     \texttt{Mistral-7B-Instruct} &     52.12 &        -1.84\\% &       4.39 &         -0.39 \\
              \texttt{Llama-2-7b} &     44.12 &        -1.76\\% &      12.09 &          3.17 \\
         \texttt{Llama-2-7b-chat} &     46.74 &        -0.48\\% &      14.05 &          1.90 \\
             \texttt{Llama-2-13b} &     52.91 &        -2.16\\% &       7.64 &          3.22 \\
        \texttt{Llama-2-13b-chat} &     52.19 &        -1.33\\% &      11.44 &          3.12 \\
                  \texttt{Yi-34B} &     73.62 &        -2.76\\% &       4.76 &          2.59 \\
             \texttt{Llama-2-70b} &     66.06 &        -2.72\\% &       5.76 &          4.20 \\
        \texttt{Llama-2-70b-chat} &     60.64 &        -2.53\\% &      11.45 &          3.39 \\
\midrule
$k_\tau$ = 0.964 &&\\
\bottomrule
\end{tabular}
}
\caption{Subject independent five-shot example results with the second prompt. (follow figure \ref{fig:subject_independent_prompts} for details). Changes in the initial prompt only result in negligible differences when compared to the first prompt in Table \ref{tab:subject_indep_fewshot1}.}
\label{tab:subject_indep_fewshot2}
\end{table}

%% file: tables/mmlu_cheating_wrong.tex
% \begin{table}[h!]
% \centering
% \resizebox{\linewidth}{!}{
% \begin{tabular}{lccccc}
% \toprule
%                                    & \multicolumn{2}{c}{One-shot} & \multicolumn{2}{c}{Five-shot} \\
% \cmidrule(lr){2-3} \cmidrule(lr){4-5}
% Model                       & Task acc & $\Delta$Acc & Task acc & $\Delta$Acc \\
% \midrule
% \texttt{phi-2}              & 33.59 & -20.88 & 13.91 & -42.86 \\
% \texttt{Yi-6B}              & 36.13 & -24.99 & 17.97 & -45.26 \\
% \texttt{Mistral-7B}         & 19.51 & -40.05 & 13.20 & -49.16 \\
% \texttt{Mistral-7B-Instruct}& 10.71 & -42.77 & 4.59  & -49.36 \\
% \texttt{Llama-2-7b}         & 24.25 & -17.56 & 23.63 & -22.25 \\
% \texttt{Llama-2-7b-chat}    & 16.24 & -30.12 & 28.11 & -19.11 \\
% \texttt{Llama-2-13b}        & 12.76 & -39.32 & 4.50  & -50.56 \\
% \texttt{Llama-2-13b-chat}   & 31.49 & -21.63 & 26.30 & -27.22 \\
% \texttt{Yi-34B}             & 32.08 & -41.30 & 37.42 & -38.96 \\
% \texttt{Llama-2-70b}        & 26.27 & -39.17 & 21.54 & -47.24 \\
% \texttt{Llama-2-70b-chat}   & 26.26 & -34.85 & 37.23 & -25.94 \\
% \midrule
% $k_\tau$         & \multicolumn{2}{c}{0.382} & \multicolumn{2}{c}{0.164} \\
% \bottomrule
% \end{tabular}
% }
% \caption{Providing the incorrect answer in-context. Performance drastically drops across the board, indicating that models are easily influenced by the answers given in-context, even when they are incorrect.}
% \label{tab:mmlu_cheating_wrong}
% \end{table}

\begin{table}[h!]
\centering
\resizebox{\linewidth}{!}{
\begin{tabular}{lccc}
\toprule

Model  & Task Acc  & Task Acc  \\
& 1-shot & 5-shot \\
\midrule
\texttt{phi-2}              & 33.59  & 13.91  \\
\texttt{Yi-6B}              & 36.13 & 17.97  \\
\texttt{Mistral-7B}         & 19.51 & 13.20  \\
\texttt{Mistral-7B-Instruct}& 10.71 & 4.59   \\
\texttt{Llama-2-7b}         & 24.25 & 23.63 \\
\texttt{Llama-2-7b-chat}    & 16.24 & 28.11 \\
\texttt{Llama-2-13b}        & 12.76 & 4.50  \\
\texttt{Llama-2-13b-chat}   & 31.49 & 26.30 \\
\texttt{Yi-34B}             & 32.08 & 37.42  \\
\texttt{Llama-2-70b}        & 26.27 & 21.54 \\
\texttt{Llama-2-70b-chat}   & 26.26 & 37.23  \\
\midrule
$k_\tau$         & 0.382 & 0.164 \\
\bottomrule
\end{tabular}
}
\caption{Providing the incorrect answer in-context. Performance drastically drops across the board, indicating that models are easily influenced by the answers given in-context, even when they are incorrect.}
\label{tab:mmlu_cheating_wrong}
\end{table}

%% file: tables/mmlu_cheating.tex
% \begin{table}[h]
% \centering
% \resizebox{\linewidth}{!}{%
% \begin{tabular}{lcccc}
% \toprule
%                                    & \multicolumn{2}{c}{One-shot} & \multicolumn{2}{c}{Five-shot} \\
% \cmidrule(lr){2-3} \cmidrule(lr){4-5}
% Model & Task Acc & $\Delta$Acc & Task Acc & $\Delta$Acc \\
% \midrule
% \texttt{phi-2} & 71.778 & 16.579 & 92.366 & 35.593 \\
% \texttt{Yi-6B} & 90.91 & 29.23 & 97.09 & 33.86 \\
% \texttt{Mistral-7B} & 97.45 & 36.85 & 98.99 & 36.63 \\
% \texttt{Mistral-7B-Instruct} & 98.64 & 45.61 & 99.25 & 45.29 \\
% \texttt{Llama-2-7b} & 61.00 & 17.68 & 63.82 & 17.94 \\
% \texttt{Llama-2-7b-chat} & 87.77 & 41.65 & 80.15 & 32.93 \\
% \texttt{Llama-2-13b} & 96.60 & 43.86 & 99.79 & 44.72 \\
% \texttt{Llama-2-13b-chat} & 87.02 & 35.11 & 92.69 & 39.17 \\
% \texttt{Yi-34B} & 99.10 & 23.87 & 98.50 & 22.12 \\
% \texttt{Llama-2-70b} & 93.45 & 25.75 & 99.09 & 30.31 \\
% \texttt{Llama-2-70b-chat} & 98.25 & 36.43 & 93.86 & 30.69 \\
% \midrule
% $k_\tau$ & \multicolumn{2}{c}{0.491} & \multicolumn{2}{c}{0.382} \\
% \bottomrule
% \end{tabular}
% }
% \caption{Results of the one-shot and five-shots MMLU in-context cheating experiment. Performance expectedly increases, indicating that models are readily able to "cheat" from the given few-shot examples in both five-shots and one-shot cases. However, no model achieved 100\% accuracy, so we encouraged the investigation of misclassified samples to validate their correctness.}
% \label{tab:mmlu_cheating_combined}
% \end{table}

\begin{table}[h!]
\centering
\resizebox{\linewidth}{!}{%
\begin{tabular}{lccc}
\toprule                         
 & Task Acc & Task Acc  \\
Model & 1-shot & 5-shot \\
\midrule
\texttt{phi-2} & 71.778 & 92.366 \\
\texttt{Yi-6B} & 90.91 & 97.09 \\
\texttt{Mistral-7B} & 97.45  & 98.99  \\
\texttt{Mistral-7B-Instruct} & 98.64 & 99.25 \\
\texttt{Llama-2-7b} & 61.00 & 63.82 \\
\texttt{Llama-2-7b-chat} & 87.77  & 80.15  \\
\texttt{Llama-2-13b} & 96.60  & 99.79  \\
\texttt{Llama-2-13b-chat} & 87.02  & 92.69 \\
\texttt{Yi-34B} & 99.10  & 98.50  \\
\texttt{Llama-2-70b} & 93.45  & 99.09 \\
\texttt{Llama-2-70b-chat} & 98.25  & 93.86 \\
\midrule
$k_\tau$ & 0.491 & 0.382 \\
\bottomrule
\end{tabular}

}
\caption{Results of the one-shot and five-shot MMLU in-context cheating experiment. Performance expectedly increases, indicating that models are readily able to "cheat" from the given few-shot examples in both five-shot and one-shot cases. However, no model achieved 100\% accuracy, so we encourage the investigation of misclassified samples to validate their correctness.}
\label{tab:mmlu_cheating_combined}
\end{table}

%% file: tables/5shot_bias_unchanged_long.tex
% makecell regex (\d+\.\d+) (\([+-]\d+\.\d+\)) -> \makecell{$1 \\\\ $2}

\begin{table}[ht]
\centering
\resizebox{\linewidth}{!}{
\begin{tabular}{lccccc}
\toprule
 & \textbf{5-shot Baseline} & \textbf{A} & \textbf{B} & \textbf{C} & \textbf{D} \\
\midrule
\texttt{phi-2} & 56.77 & \makecell{36.67 \\ \textcolor{red}{(-20.11)}} & \makecell{41.33 \\ \textcolor{red}{(-15.44)}} & \makecell{40.67 \\ \textcolor{red}{(-16.11)}} & \makecell{41.67 \\ \textcolor{red}{(-15.11)}} \\
\texttt{Yi-6B} & 63.23 & \makecell{36.67 \\ \textcolor{red}{(-26.56)}} & \makecell{36.33 \\ \textcolor{red}{(-26.89)}} & \makecell{37.67 \\ \textcolor{red}{(-25.56)}} & \makecell{39.33 \\ \textcolor{red}{(-23.89)}} \\
\texttt{Mistral-7B} & 62.36 & \makecell{34.67 \\ \textcolor{red}{(-27.70)}} & \makecell{41.33 \\ \textcolor{red}{(-21.03)}} & \makecell{43.00 \\ \textcolor{red}{(-19.36)}} & \makecell{40.33 \\ \textcolor{red}{(-22.03)}} \\
\texttt{Mistral-7B-Instruct} & 53.95 & \makecell{32.67 \\ \textcolor{red}{(-21.29)}} & \makecell{33.33 \\ \textcolor{red}{(-20.62)}} & \makecell{30.67 \\ \textcolor{red}{(-23.29)}} & \makecell{35.33 \\ \textcolor{red}{(-18.62)}} \\
\texttt{Llama-2-7b} & 45.88 & \makecell{22.00 \\ \textcolor{red}{(-23.88)}} & \makecell{31.00 \\ \textcolor{red}{(-14.88)}} & \makecell{30.67 \\ \textcolor{red}{(-15.22)}} & \makecell{34.33 \\ \textcolor{red}{(-11.55)}} \\
\texttt{Llama-2-7b-chat} & 47.22 & \makecell{31.00 \\ \textcolor{red}{(-16.22)}} & \makecell{30.67 \\ \textcolor{red}{(-16.56)}} & \makecell{28.67 \\ \textcolor{red}{(-18.56)}} & \makecell{31.00 \\ \textcolor{red}{(-16.22)}} \\
\texttt{Llama-2-13b} & 55.06 & \makecell{35.33 \\ \textcolor{red}{(-19.73)}} & \makecell{36.33 \\ \textcolor{red}{(-18.73)}} & \makecell{37.67 \\ \textcolor{red}{(-17.40)}} & \makecell{32.67 \\ \textcolor{red}{(-22.40)}} \\
\texttt{Llama-2-13b-chat} & 53.53 & \makecell{31.67 \\ \textcolor{red}{(-21.86)}} & \makecell{33.00 \\ \textcolor{red}{(-20.53)}} & \makecell{34.67 \\ \textcolor{red}{(-18.86)}} & \makecell{33.67 \\ \textcolor{red}{(-19.86)}} \\
\texttt{Yi-34B} & 76.39 & \makecell{49.67 \\ \textcolor{red}{(-26.72)}} & \makecell{49.33 \\ \textcolor{red}{(-27.05)}} & \makecell{50.33 \\ \textcolor{red}{(-26.05)}} & \makecell{48.67 \\ \textcolor{red}{(-27.72)}} \\
\texttt{Llama-2-70b} & 68.78 & \makecell{42.67 \\ \textcolor{red}{(-26.11)}} & \makecell{44.67 \\ \textcolor{red}{(-24.11)}} & \makecell{43.33 \\ \textcolor{red}{(-25.45)}} & \makecell{44.33 \\ \textcolor{red}{(-24.45)}} \\
\texttt{Llama-2-70b-chat} & 63.17 & \makecell{40.33 \\ \textcolor{red}{(-22.84)}} & \makecell{42.33 \\ \textcolor{red}{(-20.84)}} & \makecell{42.00 \\ \textcolor{red}{(-21.17)}} & \makecell{41.33 \\ \textcolor{red}{(-21.84)}} \\

\midrule
\text{$k_\tau$} & - & 0.855 & 0.818 & 0.782 & 0.636 \\% \texttt{K_tau} & 61.110 &
\bottomrule
\end{tabular}
}

\caption{Results of fixing the five-shot example answers to positions A/B/C/D, averaged over the three selected subjects. We can see that performance drops across the board, suggesting that models get confused when there is a clear pattern in the correct answers of the few-shot examples.}
\label{tab:five_shot_bias_unchanged_long}
\end{table}

%% file: tables/mmlu_test_stats.tex
\begin{table}[ht!]

\resizebox{\linewidth}{!}{
\begin{tabular}{lccccc}
    \hline
        subject & choice\_A & choice\_B & choice\_C & choice\_D & total \\ \hline
        abstract\_algebra & 22 & 26 & 31 & 21 & 100 \\
        anatomy & 25 & 34 & 45 & 31 & 135 \\
        astronomy & 27 & 28 & 46 & 51 & 152 \\
        business\_ethics & 30 & 26 & 23 & 21 & 100 \\
        clinical\_knowledge & 57 & 71 & 58 & 79 & 265 \\
        college\_biology & 37 & 32 & 37 & 38 & 144 \\
        college\_chemistry & 20 & 21 & 18 & 41 & 100 \\
        college\_computer\_science & 26 & 15 & 26 & 33 & 100 \\
        college\_mathematics & 21 & 23 & 25 & 31 & 100 \\
        college\_medicine & 36 & 36 & 43 & 58 & 173 \\
        college\_physics & 22 & 20 & 22 & 38 & 102 \\
        computer\_security & 28 & 24 & 30 & 18 & 100 \\
        conceptual\_physics & 62 & 76 & 48 & 49 & 235 \\
        econometrics & 27 & 32 & 28 & 27 & 114 \\
        electrical\_engineering & 35 & 32 & 43 & 35 & 145 \\
        elementary\_mathematics & 79 & 97 & 101 & 101 & 378 \\
        formal\_logic & 36 & 25 & 19 & 46 & 126 \\
        global\_facts & 18 & 31 & 33 & 18 & 100 \\
        high\_school\_biology & 55 & 79 & 78 & 98 & 310 \\
        high\_school\_chemistry & 31 & 55 & 60 & 57 & 203 \\
        high\_school\_computer\_science & 25 & 23 & 33 & 19 & 100 \\
        high\_school\_european\_history & 36 & 40 & 47 & 42 & 165 \\
        high\_school\_geography & 35 & 43 & 50 & 70 & 198 \\
        high\_school\_government\_and\_politics & 38 & 40 & 44 & 71 & 193 \\
        high\_school\_macroeconomics & 79 & 86 & 83 & 142 & 390 \\
        high\_school\_mathematics & 57 & 71 & 71 & 71 & 270 \\
        high\_school\_microeconomics & 50 & 55 & 50 & 83 & 238 \\
        high\_school\_physics & 30 & 30 & 41 & 50 & 151 \\
        high\_school\_psychology & 105 & 129 & 121 & 190 & 545 \\
        high\_school\_statistics & 33 & 35 & 46 & 102 & 216 \\
        high\_school\_us\_history & 51 & 48 & 53 & 52 & 204 \\
        high\_school\_world\_history & 64 & 62 & 63 & 48 & 237 \\
        human\_aging & 70 & 84 & 45 & 24 & 223 \\
        human\_sexuality & 34 & 30 & 30 & 37 & 131 \\
        international\_law & 29 & 30 & 45 & 17 & 121 \\
        jurisprudence & 28 & 32 & 25 & 23 & 108 \\
        logical\_fallacies & 36 & 40 & 49 & 38 & 163 \\
        machine\_learning & 35 & 32 & 27 & 18 & 112 \\
        management & 18 & 26 & 20 & 39 & 103 \\
        marketing & 68 & 60 & 60 & 46 & 234 \\
        medical\_genetics & 30 & 26 & 20 & 24 & 100 \\
        miscellaneous & 186 & 225 & 212 & 160 & 783 \\
        moral\_disputes & 86 & 85 & 101 & 74 & 346 \\
        moral\_scenarios & 213 & 217 & 221 & 244 & 895 \\
        nutrition & 69 & 70 & 77 & 90 & 306 \\
        philosophy & 58 & 85 & 93 & 75 & 311 \\
        prehistory & 70 & 86 & 95 & 73 & 324 \\
        professional\_accounting & 66 & 72 & 76 & 68 & 282 \\
        professional\_law & 377 & 367 & 415 & 375 & 1534 \\
        professional\_medicine & 50 & 55 & 45 & 122 & 272 \\
        professional\_psychology & 153 & 157 & 169 & 133 & 612 \\
        public\_relations & 24 & 38 & 23 & 25 & 110 \\
        security\_studies & 46 & 42 & 59 & 98 & 245 \\
        sociology & 49 & 48 & 50 & 54 & 201 \\
        us\_foreign\_policy & 28 & 21 & 25 & 26 & 100 \\
        virology & 47 & 53 & 34 & 32 & 166 \\
        world\_religions & 55 & 36 & 50 & 30 & 171 \\ \hline
        total & 3222 & 3462 & 3582 & 3776 & 14042 \\ \hline
    \end{tabular}
    }
        \caption{The MMLU subjects statistics in the test split.}
\label{tab:mmlu_test_stats}

\end{table}

%% file: tables/mmlu_dev_stats.tex
\begin{table}[h!]
\resizebox{\linewidth}{!}{
\begin{tabular}{lccccc}
    \hline
        subject & choice\_A & choice\_B & choice\_C & choice\_D & total \\  \hline
        abstract\_algebra & 2 & 2 & 1 & 0 & 5 \\
        anatomy & 0 & 1 & 2 & 2 & 5 \\
        astronomy & 3 & 0 & 0 & 2 & 5 \\
        business\_ethics & 1 & 1 & 1 & 2 & 5 \\
        clinical\_knowledge & 1 & 2 & 2 & 0 & 5 \\
        college\_biology & 0 & 3 & 1 & 1 & 5 \\
        college\_chemistry & 2 & 0 & 1 & 2 & 5 \\
        college\_computer\_science & 0 & 2 & 0 & 3 & 5 \\
        college\_mathematics & 0 & 2 & 1 & 2 & 5 \\
        college\_medicine & 2 & 1 & 1 & 1 & 5 \\
        college\_physics & 3 & 1 & 0 & 1 & 5 \\
        computer\_security & 1 & 1 & 1 & 2 & 5 \\
        conceptual\_physics & 2 & 1 & 2 & 0 & 5 \\
        econometrics & 1 & 0 & 3 & 1 & 5 \\
        electrical\_engineering & 1 & 2 & 1 & 1 & 5 \\
        elementary\_mathematics & 1 & 4 & 0 & 0 & 5 \\
        formal\_logic & 0 & 1 & 3 & 1 & 5 \\
        global\_facts & 2 & 3 & 0 & 0 & 5 \\
        high\_school\_biology & 1 & 0 & 1 & 3 & 5 \\
        high\_school\_chemistry & 1 & 0 & 3 & 1 & 5 \\
        high\_school\_computer\_science & 0 & 1 & 3 & 1 & 5 \\
        high\_school\_european\_history & 2 & 1 & 1 & 1 & 5 \\
        high\_school\_geography & 1 & 2 & 1 & 1 & 5 \\
        high\_school\_government\_and\_politics & 1 & 0 & 2 & 2 & 5 \\
        high\_school\_macroeconomics & 0 & 0 & 3 & 2 & 5 \\
        high\_school\_mathematics & 0 & 1 & 2 & 2 & 5 \\
        high\_school\_microeconomics & 0 & 1 & 2 & 2 & 5 \\
        high\_school\_physics & 0 & 2 & 0 & 3 & 5 \\
        high\_school\_psychology & 1 & 2 & 1 & 1 & 5 \\
        high\_school\_statistics & 0 & 0 & 1 & 4 & 5 \\
        high\_school\_us\_history & 0 & 2 & 2 & 1 & 5 \\
        high\_school\_world\_history & 1 & 3 & 0 & 1 & 5 \\
        human\_aging & 1 & 2 & 2 & 0 & 5 \\
        human\_sexuality & 3 & 1 & 1 & 0 & 5 \\
        international\_law & 2 & 2 & 1 & 0 & 5 \\
        jurisprudence & 2 & 0 & 0 & 3 & 5 \\
        logical\_fallacies & 0 & 1 & 2 & 2 & 5 \\
        machine\_learning & 1 & 1 & 2 & 1 & 5 \\
        management & 3 & 0 & 1 & 1 & 5 \\
        marketing & 1 & 1 & 0 & 3 & 5 \\
        medical\_genetics & 3 & 0 & 1 & 1 & 5 \\
        miscellaneous & 1 & 2 & 2 & 0 & 5 \\
        moral\_disputes & 3 & 1 & 0 & 1 & 5 \\
        moral\_scenarios & 1 & 1 & 2 & 1 & 5 \\
        nutrition & 1 & 1 & 2 & 1 & 5 \\
        philosophy & 1 & 0 & 3 & 1 & 5 \\
        prehistory & 2 & 1 & 1 & 1 & 5 \\
        professional\_accounting & 2 & 2 & 0 & 1 & 5 \\
        professional\_law & 2 & 2 & 0 & 1 & 5 \\
        professional\_medicine & 0 & 0 & 1 & 4 & 5 \\
        professional\_psychology & 2 & 0 & 0 & 3 & 5 \\
        public\_relations & 1 & 0 & 2 & 2 & 5 \\
        security\_studies & 0 & 2 & 2 & 1 & 5 \\
        sociology & 1 & 3 & 1 & 0 & 5 \\
        us\_foreign\_policy & 1 & 0 & 2 & 2 & 5 \\
        virology & 2 & 1 & 1 & 1 & 5 \\
        world\_religions & 1 & 3 & 0 & 1 & 5 \\  \hline
        total  & 67 & 69 & 71 & 78 & 285 \\ \hline
    \end{tabular}
    }
    \caption{The MMLU subjects statistics in the dev split, which is a fixed set of questions per subject used in fewshots evaluation.}
\label{tab:mmlu_dev_stats}
\end{table}

%% file: main.bbl
\begin{thebibliography}{45}
\expandafter\ifx\csname natexlab\endcsname\relax\def\natexlab#1{#1}\fi

\bibitem[{Anil et~al.(2023)Anil, Dai, Firat, Johnson, Lepikhin, Passos, Shakeri, Taropa, Bailey, Chen, Chu, Clark, Shafey, Huang, Meier-Hellstern, Mishra, Moreira, Omernick, Robinson, Ruder, Tay, Xiao, Xu, Zhang, Abrego, Ahn, Austin, Barham, Botha, Bradbury, Brahma, Brooks, Catasta, Cheng, Cherry, Choquette-Choo, Chowdhery, Crepy, Dave, Dehghani, Dev, Devlin, Díaz, Du, Dyer, Feinberg, Feng, Fienber, Freitag, Garcia, Gehrmann, Gonzalez, Gur-Ari, Hand, Hashemi, Hou, Howland, Hu, Hui, Hurwitz, Isard, Ittycheriah, Jagielski, Jia, Kenealy, Krikun, Kudugunta, Lan, Lee, Lee, Li, Li, Li, Li, Li, Lim, Lin, Liu, Liu, Maggioni, Mahendru, Maynez, Misra, Moussalem, Nado, Nham, Ni, Nystrom, Parrish, Pellat, Polacek, Polozov, Pope, Qiao, Reif, Richter, Riley, Ros, Roy, Saeta, Samuel, Shelby, Slone, Smilkov, So, Sohn, Tokumine, Valter, Vasudevan, Vodrahalli, Wang, Wang, Wang, Wang, Wieting, Wu, Xu, Xu, Xue, Yin, Yu, Zhang, Zheng, Zheng, Zhou, Zhou, Petrov, and Wu}]{palm2}
Rohan Anil, Andrew~M. Dai, Orhan Firat, Melvin Johnson, Dmitry Lepikhin, Alexandre Passos, Siamak Shakeri, Emanuel Taropa, Paige Bailey, Zhifeng Chen, Eric Chu, Jonathan~H. Clark, Laurent~El Shafey, Yanping Huang, Kathy Meier-Hellstern, Gaurav Mishra, Erica Moreira, Mark Omernick, Kevin Robinson, Sebastian Ruder, Yi~Tay, Kefan Xiao, Yuanzhong Xu, Yujing Zhang, Gustavo~Hernandez Abrego, Junwhan Ahn, Jacob Austin, Paul Barham, Jan Botha, James Bradbury, Siddhartha Brahma, Kevin Brooks, Michele Catasta, Yong Cheng, Colin Cherry, Christopher~A. Choquette-Choo, Aakanksha Chowdhery, Clément Crepy, Shachi Dave, Mostafa Dehghani, Sunipa Dev, Jacob Devlin, Mark Díaz, Nan Du, Ethan Dyer, Vlad Feinberg, Fangxiaoyu Feng, Vlad Fienber, Markus Freitag, Xavier Garcia, Sebastian Gehrmann, Lucas Gonzalez, Guy Gur-Ari, Steven Hand, Hadi Hashemi, Le~Hou, Joshua Howland, Andrea Hu, Jeffrey Hui, Jeremy Hurwitz, Michael Isard, Abe Ittycheriah, Matthew Jagielski, Wenhao Jia, Kathleen Kenealy, Maxim Krikun, Sneha Kudugunta, Chang
  Lan, Katherine Lee, Benjamin Lee, Eric Li, Music Li, Wei Li, YaGuang Li, Jian Li, Hyeontaek Lim, Hanzhao Lin, Zhongtao Liu, Frederick Liu, Marcello Maggioni, Aroma Mahendru, Joshua Maynez, Vedant Misra, Maysam Moussalem, Zachary Nado, John Nham, Eric Ni, Andrew Nystrom, Alicia Parrish, Marie Pellat, Martin Polacek, Alex Polozov, Reiner Pope, Siyuan Qiao, Emily Reif, Bryan Richter, Parker Riley, Alex~Castro Ros, Aurko Roy, Brennan Saeta, Rajkumar Samuel, Renee Shelby, Ambrose Slone, Daniel Smilkov, David~R. So, Daniel Sohn, Simon Tokumine, Dasha Valter, Vijay Vasudevan, Kiran Vodrahalli, Xuezhi Wang, Pidong Wang, Zirui Wang, Tao Wang, John Wieting, Yuhuai Wu, Kelvin Xu, Yunhan Xu, Linting Xue, Pengcheng Yin, Jiahui Yu, Qiao Zhang, Steven Zheng, Ce~Zheng, Weikang Zhou, Denny Zhou, Slav Petrov, and Yonghui Wu. 2023.
\newblock \href {http://arxiv.org/abs/2305.10403} {Palm 2 technical report}.

\bibitem[{Anthropic(2023)}]{claude2}
Anthropic. 2023.
\newblock \href {https://www-files.anthropic.com/production/images/Model-Card-Claude-2.pdf} {Anthropic. model card and evaluations for claude models.}

\bibitem[{Beeching et~al.(2023)Beeching, Fourrier, Habib, Han, Lambert, Rajani, Sanseviero, Tunstall, and Wolf}]{open-llm-leaderboard}
Edward Beeching, Clémentine Fourrier, Nathan Habib, Sheon Han, Nathan Lambert, Nazneen Rajani, Omar Sanseviero, Lewis Tunstall, and Thomas Wolf. 2023.
\newblock Open llm leaderboard.
\newblock \url{https://huggingface.co/spaces/HuggingFaceH4/open_llm_leaderboard}.

\bibitem[{Chang et~al.(2023)Chang, Wang, Wang, Wu, Zhu, Chen, Yang, Yi, Wang, Wang et~al.}]{chang2023survey}
Yupeng Chang, Xu~Wang, Jindong Wang, Yuan Wu, Kaijie Zhu, Hao Chen, Linyi Yang, Xiaoyuan Yi, Cunxiang Wang, Yidong Wang, et~al. 2023.
\newblock A survey on evaluation of large language models.
\newblock \emph{arXiv preprint arXiv:2307.03109}.

\bibitem[{Chowdhery et~al.(2022)Chowdhery, Narang, Devlin, Bosma, Mishra, Roberts, Barham, Chung, Sutton, Gehrmann et~al.}]{chowdhery2022palm}
Aakanksha Chowdhery, Sharan Narang, Jacob Devlin, Maarten Bosma, Gaurav Mishra, Adam Roberts, Paul Barham, Hyung~Won Chung, Charles Sutton, Sebastian Gehrmann, et~al. 2022.
\newblock Palm: Scaling language modeling with pathways.
\newblock \emph{arXiv preprint arXiv:2204.02311}.

\bibitem[{Clark et~al.(2018{\natexlab{a}})Clark, Cowhey, Etzioni, Khot, Sabharwal, Schoenick, and Tafjord}]{Clark2018ThinkYH}
Peter Clark, Isaac Cowhey, Oren Etzioni, Tushar Khot, Ashish Sabharwal, Carissa Schoenick, and Oyvind Tafjord. 2018{\natexlab{a}}.
\newblock \href {https://api.semanticscholar.org/CorpusID:3922816} {Think you have solved question answering? try arc, the ai2 reasoning challenge}.
\newblock \emph{ArXiv}, abs/1803.05457.

\bibitem[{Clark et~al.(2018{\natexlab{b}})Clark, Cowhey, Etzioni, Khot, Sabharwal, Schoenick, and Tafjord}]{clark2018thinkarc}
Peter Clark, Isaac Cowhey, Oren Etzioni, Tushar Khot, Ashish Sabharwal, Carissa Schoenick, and Oyvind Tafjord. 2018{\natexlab{b}}.
\newblock Think you have solved question answering? try arc, the ai2 reasoning challenge.
\newblock \emph{arXiv preprint arXiv:1803.05457}.

\bibitem[{Contributors(2023)}]{2023opencompass}
OpenCompass Contributors. 2023.
\newblock Opencompass: A universal evaluation platform for foundation models.
\newblock \url{https://github.com/open-compass/opencompass}.

\bibitem[{Deepmind(2023)}]{gemini}
Google Deepmind. 2023.
\newblock Gemini: A family of highly capable multimodal models.

\bibitem[{Dehghani et~al.(2021)Dehghani, Tay, Gritsenko, Zhao, Houlsby, Diaz, Metzler, and Vinyals}]{dehghani2021benchmark}
Mostafa Dehghani, Yi~Tay, Alexey~A. Gritsenko, Zhe Zhao, Neil Houlsby, Fernando Diaz, Donald Metzler, and Oriol Vinyals. 2021.
\newblock \href {http://arxiv.org/abs/2107.07002} {The benchmark lottery}.

\bibitem[{Deng et~al.(2023)Deng, Zhao, Tang, Gerstein, and Cohan}]{deng2023benchmark}
Chunyuan Deng, Yilun Zhao, Xiangru Tang, Mark Gerstein, and Arman Cohan. 2023.
\newblock Benchmark probing: Investigating data leakage in large language models.
\newblock In \emph{NeurIPS 2023 Workshop on Backdoors in Deep Learning-The Good, the Bad, and the Ugly}.

\bibitem[{Ethayarajh and Jurafsky(2021)}]{ethayarajh2021utility}
Kawin Ethayarajh and Dan Jurafsky. 2021.
\newblock \href {http://arxiv.org/abs/2009.13888} {Utility is in the eye of the user: A critique of nlp leaderboards}.

\bibitem[{Gao et~al.(2023)Gao, Tow, Abbasi, Biderman, Black, DiPofi, Foster, Golding, Hsu, Le~Noac'h, Li, McDonell, Muennighoff, Ociepa, Phang, Reynolds, Schoelkopf, Skowron, Sutawika, Tang, Thite, Wang, Wang, and Zou}]{eval-harness}
Leo Gao, Jonathan Tow, Baber Abbasi, Stella Biderman, Sid Black, Anthony DiPofi, Charles Foster, Laurence Golding, Jeffrey Hsu, Alain Le~Noac'h, Haonan Li, Kyle McDonell, Niklas Muennighoff, Chris Ociepa, Jason Phang, Laria Reynolds, Hailey Schoelkopf, Aviya Skowron, Lintang Sutawika, Eric Tang, Anish Thite, Ben Wang, Kevin Wang, and Andy Zou. 2023.
\newblock \href {https://doi.org/10.5281/zenodo.10256836} {A framework for few-shot language model evaluation}.

\bibitem[{Hendrycks et~al.(2020)Hendrycks, Burns, Basart, Zou, Mazeika, Song, and Steinhardt}]{hendrycksmeasuring}
Dan Hendrycks, Collin Burns, Steven Basart, Andy Zou, Mantas Mazeika, Dawn Song, and Jacob Steinhardt. 2020.
\newblock Measuring massive multitask language understanding.
\newblock In \emph{International Conference on Learning Representations}.

\bibitem[{Jiang et~al.(2023)Jiang, Sablayrolles, Mensch, Bamford, Chaplot, Casas, Bressand, Lengyel, Lample, Saulnier et~al.}]{jiang2023mistral}
Albert~Q Jiang, Alexandre Sablayrolles, Arthur Mensch, Chris Bamford, Devendra~Singh Chaplot, Diego de~las Casas, Florian Bressand, Gianna Lengyel, Guillaume Lample, Lucile Saulnier, et~al. 2023.
\newblock Mistral 7b.
\newblock \emph{arXiv preprint arXiv:2310.06825}.

\bibitem[{Kendall(1938)}]{kendall1938new}
Maurice~G Kendall. 1938.
\newblock A new measure of rank correlation.
\newblock \emph{Biometrika}, 30(1/2):81--93.

\bibitem[{Laskar et~al.(2023)Laskar, Bari, Rahman, Bhuiyan, Joty, and Huang}]{laskar2023systematic}
Md~Tahmid~Rahman Laskar, M~Saiful Bari, Mizanur Rahman, Md~Amran~Hossen Bhuiyan, Shafiq Joty, and Jimmy~Xiangji Huang. 2023.
\newblock A systematic study and comprehensive evaluation of chatgpt on benchmark datasets.
\newblock \emph{arXiv preprint arXiv:2305.18486}.

\bibitem[{Liang et~al.(2023)Liang, Bommasani, Lee, Tsipras, Soylu, Yasunaga, Zhang, Narayanan, Wu, Kumar, Newman, Yuan, Yan, Zhang, Cosgrove, Manning, Ré, Acosta-Navas, Hudson, Zelikman, Durmus, Ladhak, Rong, Ren, Yao, Wang, Santhanam, Orr, Zheng, Yuksekgonul, Suzgun, Kim, Guha, Chatterji, Khattab, Henderson, Huang, Chi, Xie, Santurkar, Ganguli, Hashimoto, Icard, Zhang, Chaudhary, Wang, Li, Mai, Zhang, and Koreeda}]{liang2023holistic}
Percy Liang, Rishi Bommasani, Tony Lee, Dimitris Tsipras, Dilara Soylu, Michihiro Yasunaga, Yian Zhang, Deepak Narayanan, Yuhuai Wu, Ananya Kumar, Benjamin Newman, Binhang Yuan, Bobby Yan, Ce~Zhang, Christian Cosgrove, Christopher~D. Manning, Christopher Ré, Diana Acosta-Navas, Drew~A. Hudson, Eric Zelikman, Esin Durmus, Faisal Ladhak, Frieda Rong, Hongyu Ren, Huaxiu Yao, Jue Wang, Keshav Santhanam, Laurel Orr, Lucia Zheng, Mert Yuksekgonul, Mirac Suzgun, Nathan Kim, Neel Guha, Niladri Chatterji, Omar Khattab, Peter Henderson, Qian Huang, Ryan Chi, Sang~Michael Xie, Shibani Santurkar, Surya Ganguli, Tatsunori Hashimoto, Thomas Icard, Tianyi Zhang, Vishrav Chaudhary, William Wang, Xuechen Li, Yifan Mai, Yuhui Zhang, and Yuta Koreeda. 2023.
\newblock \href {http://arxiv.org/abs/2211.09110} {Holistic evaluation of language models}.

\bibitem[{Lions et~al.(2021)Lions, Monsalve, Dartnell, Godoy, C{\'o}rdova, Jim{\'e}nez, Blanco, Ortega, and Lemari{\'e}}]{lions2021position}
S{\'e}verin Lions, Carlos Monsalve, Pablo Dartnell, Mar{\'\i}a~In{\'e}s Godoy, Nora C{\'o}rdova, Daniela Jim{\'e}nez, Mar{\'\i}a~Paz Blanco, Gabriel Ortega, and Julie Lemari{\'e}. 2021.
\newblock The position of distractors in multiple-choice test items: The strongest precede the weakest.
\newblock In \emph{Frontiers in Education}, volume~6, page 731763. Frontiers.

\bibitem[{Lu et~al.(2022)Lu, Bartolo, Moore, Riedel, and Stenetorp}]{lu-etal-2022-fantastically}
Yao Lu, Max Bartolo, Alastair Moore, Sebastian Riedel, and Pontus Stenetorp. 2022.
\newblock \href {https://doi.org/10.18653/v1/2022.acl-long.556} {Fantastically ordered prompts and where to find them: Overcoming few-shot prompt order sensitivity}.
\newblock In \emph{Proceedings of the 60th Annual Meeting of the Association for Computational Linguistics (Volume 1: Long Papers)}, pages 8086--8098, Dublin, Ireland. Association for Computational Linguistics.

\bibitem[{Min et~al.(2022)Min, Lyu, Holtzman, Artetxe, Lewis, Hajishirzi, and Zettlemoyer}]{min-etal-2022-rethinking}
Sewon Min, Xinxi Lyu, Ari Holtzman, Mikel Artetxe, Mike Lewis, Hannaneh Hajishirzi, and Luke Zettlemoyer. 2022.
\newblock \href {https://doi.org/10.18653/v1/2022.emnlp-main.759} {Rethinking the role of demonstrations: What makes in-context learning work?}
\newblock In \emph{Proceedings of the 2022 Conference on Empirical Methods in Natural Language Processing}, pages 11048--11064, Abu Dhabi, United Arab Emirates. Association for Computational Linguistics.

\bibitem[{Mishra et~al.(2022)Mishra, Khashabi, Baral, and Hajishirzi}]{mishra2022crosstask}
Swaroop Mishra, Daniel Khashabi, Chitta Baral, and Hannaneh Hajishirzi. 2022.
\newblock \href {http://arxiv.org/abs/2104.08773} {Cross-task generalization via natural language crowdsourcing instructions}.

\bibitem[{Mizrahi et~al.(2024)Mizrahi, Kaplan, Malkin, Dror, Shahaf, and Stanovsky}]{mizrahi2024state}
Moran Mizrahi, Guy Kaplan, Dan Malkin, Rotem Dror, Dafna Shahaf, and Gabriel Stanovsky. 2024.
\newblock \href {http://arxiv.org/abs/2401.00595} {State of what art? a call for multi-prompt llm evaluation}.

\bibitem[{Nie et~al.(2019)Nie, Williams, Dinan, Bansal, Weston, and Kiela}]{nie2019adversarial}
Yixin Nie, Adina Williams, Emily Dinan, Mohit Bansal, Jason Weston, and Douwe Kiela. 2019.
\newblock Adversarial nli: A new benchmark for natural language understanding.
\newblock \emph{arXiv preprint arXiv:1910.14599}.

\bibitem[{OpenAI(2023)}]{openai2023gpt4}
OpenAI. 2023.
\newblock \href {http://arxiv.org/abs/2303.08774} {Gpt-4 technical report}.

\bibitem[{Pezeshkpour and Hruschka(2023)}]{pezeshkpour2023large}
Pouya Pezeshkpour and Estevam Hruschka. 2023.
\newblock \href {http://arxiv.org/abs/2308.11483} {Large language models sensitivity to the order of options in multiple-choice questions}.

\bibitem[{Raffel et~al.(2020)Raffel, Shazeer, Roberts, Lee, Narang, Matena, Zhou, Li, Liu et~al.}]{t5}
Colin Raffel, Noam Shazeer, Adam Roberts, Katherine Lee, Sharan Narang, Michael Matena, Yanqi Zhou, Wei Li, Peter~J Liu, et~al. 2020.
\newblock Exploring the limits of transfer learning with a unified text-to-text transformer.
\newblock \emph{J. Mach. Learn. Res.}, 21(140):1--67.

\bibitem[{Robinson et~al.(2023)Robinson, Rytting, and Wingate}]{robinson2023leveraging}
Joshua Robinson, Christopher~Michael Rytting, and David Wingate. 2023.
\newblock \href {http://arxiv.org/abs/2210.12353} {Leveraging large language models for multiple choice question answering}.

\bibitem[{Saha et~al.(2018)Saha, Pahuja, Khapra, Sankaranarayanan, and Chandar}]{saha2018complex}
Amrita Saha, Vardaan Pahuja, Mitesh~M. Khapra, Karthik Sankaranarayanan, and Sarath Chandar. 2018.
\newblock \href {http://arxiv.org/abs/1801.10314} {Complex sequential question answering: Towards learning to converse over linked question answer pairs with a knowledge graph}.

\bibitem[{Sanh et~al.(2021)Sanh, Webson, Raffel, Bach, Sutawika, Alyafeai, Chaffin, Stiegler, Scao, Raja et~al.}]{T0}
Victor Sanh, Albert Webson, Colin Raffel, Stephen~H Bach, Lintang Sutawika, Zaid Alyafeai, Antoine Chaffin, Arnaud Stiegler, Teven~Le Scao, Arun Raja, et~al. 2021.
\newblock \href {https://doi.org/10.48550/ARXIV.2110.08207} {Multitask prompted training enables zero-shot task generalization}.
\newblock \emph{arXiv preprint arXiv:2110.08207}.

\bibitem[{Sclar et~al.(2023)Sclar, Choi, Tsvetkov, and Suhr}]{sclar2023quantifying}
Melanie Sclar, Yejin Choi, Yulia Tsvetkov, and Alane Suhr. 2023.
\newblock \href {http://arxiv.org/abs/2310.11324} {Quantifying language models' sensitivity to spurious features in prompt design or: How i learned to start worrying about prompt formatting}.

\bibitem[{Shen et~al.(2023)Shen, Cheng, You, and Bing}]{shen2023large}
Chenhui Shen, Liying Cheng, Yang You, and Lidong Bing. 2023.
\newblock Are large language models good evaluators for abstractive summarization?
\newblock \emph{arXiv preprint arXiv:2305.13091}.

\bibitem[{Soltan et~al.(2022)Soltan, Ananthakrishnan, FitzGerald, Gupta, Hamza, Khan, Peris, Rawls, Rosenbaum, Rumshisky et~al.}]{soltan2022alexatm}
Saleh Soltan, Shankar Ananthakrishnan, Jack FitzGerald, Rahul Gupta, Wael Hamza, Haidar Khan, Charith Peris, Stephen Rawls, Andy Rosenbaum, Anna Rumshisky, et~al. 2022.
\newblock Alexatm 20b: Few-shot learning using a large-scale multilingual seq2seq model.
\newblock \emph{arXiv preprint arXiv:2208.01448}.

\bibitem[{Suzgun et~al.(2022)Suzgun, Scales, Sch{\"a}rli, Gehrmann, Tay, Chung, Chowdhery, Le, Chi, Zhou et~al.}]{suzgun2022challenging}
Mirac Suzgun, Nathan Scales, Nathanael Sch{\"a}rli, Sebastian Gehrmann, Yi~Tay, Hyung~Won Chung, Aakanksha Chowdhery, Quoc~V Le, Ed~H Chi, Denny Zhou, et~al. 2022.
\newblock Challenging big-bench tasks and whether chain-of-thought can solve them.
\newblock \emph{arXiv preprint arXiv:2210.09261}.

\bibitem[{Touvron et~al.(2023)Touvron, Martin, Stone, Albert, Almahairi, Babaei, Bashlykov, Batra, Bhargava, Bhosale, Bikel, Blecher, Ferrer, Chen, Cucurull, Esiobu, Fernandes, Fu, Fu, Fuller, Gao, Goswami, Goyal, Hartshorn, Hosseini, Hou, Inan, Kardas, Kerkez, Khabsa, Kloumann, Korenev, Koura, Lachaux, Lavril, Lee, Liskovich, Lu, Mao, Martinet, Mihaylov, Mishra, Molybog, Nie, Poulton, Reizenstein, Rungta, Saladi, Schelten, Silva, Smith, Subramanian, Tan, Tang, Taylor, Williams, Kuan, Xu, Yan, Zarov, Zhang, Fan, Kambadur, Narang, Rodriguez, Stojnic, Edunov, and Scialom}]{touvron2023llama2}
Hugo Touvron, Louis Martin, Kevin Stone, Peter Albert, Amjad Almahairi, Yasmine Babaei, Nikolay Bashlykov, Soumya Batra, Prajjwal Bhargava, Shruti Bhosale, Dan Bikel, Lukas Blecher, Cristian~Canton Ferrer, Moya Chen, Guillem Cucurull, David Esiobu, Jude Fernandes, Jeremy Fu, Wenyin Fu, Brian Fuller, Cynthia Gao, Vedanuj Goswami, Naman Goyal, Anthony Hartshorn, Saghar Hosseini, Rui Hou, Hakan Inan, Marcin Kardas, Viktor Kerkez, Madian Khabsa, Isabel Kloumann, Artem Korenev, Punit~Singh Koura, Marie-Anne Lachaux, Thibaut Lavril, Jenya Lee, Diana Liskovich, Yinghai Lu, Yuning Mao, Xavier Martinet, Todor Mihaylov, Pushkar Mishra, Igor Molybog, Yixin Nie, Andrew Poulton, Jeremy Reizenstein, Rashi Rungta, Kalyan Saladi, Alan Schelten, Ruan Silva, Eric~Michael Smith, Ranjan Subramanian, Xiaoqing~Ellen Tan, Binh Tang, Ross Taylor, Adina Williams, Jian~Xiang Kuan, Puxin Xu, Zheng Yan, Iliyan Zarov, Yuchen Zhang, Angela Fan, Melanie Kambadur, Sharan Narang, Aurelien Rodriguez, Robert Stojnic, Sergey Edunov, and Thomas
  Scialom. 2023.
\newblock \href {http://arxiv.org/abs/2307.09288} {Llama 2: Open foundation and fine-tuned chat models}.

\bibitem[{Wang et~al.(2019)Wang, Pruksachatkun, Nangia, Singh, Michael, Hill, Levy, and Bowman}]{super_glue}
Alex Wang, Yada Pruksachatkun, Nikita Nangia, Amanpreet Singh, Julian Michael, Felix Hill, Omer Levy, and Samuel Bowman. 2019.
\newblock \href {https://proceedings.neurips.cc/paper/2019/hash/4496bf24afe7fab6f046bf4923da8de6-Abstract.html} {Superglue: A stickier benchmark for general-purpose language understanding systems}.
\newblock \emph{Advances in neural information processing systems}, 32.

\bibitem[{Wang et~al.(2018)Wang, Singh, Michael, Hill, Levy, and Bowman}]{wang2018glue}
Alex Wang, Amanpreet Singh, Julian Michael, Felix Hill, Omer Levy, and Samuel~R Bowman. 2018.
\newblock Glue: A multi-task benchmark and analysis platform for natural language understanding.
\newblock \emph{arXiv preprint arXiv:1804.07461}.

\bibitem[{Wang et~al.(2023)Wang, Feng, Wang, Shi, Balachandran, He, and Tsvetkov}]{wang2023resolving}
Yike Wang, Shangbin Feng, Heng Wang, Weijia Shi, Vidhisha Balachandran, Tianxing He, and Yulia Tsvetkov. 2023.
\newblock \href {http://arxiv.org/abs/2310.00935} {Resolving knowledge conflicts in large language models}.

\bibitem[{Weber et~al.(2023)Weber, Bruni, and Hupkes}]{weber-etal-2023-mind}
Lucas Weber, Elia Bruni, and Dieuwke Hupkes. 2023.
\newblock \href {https://doi.org/10.18653/v1/2023.conll-1.20} {Mind the instructions: a holistic evaluation of consistency and interactions in prompt-based learning}.
\newblock In \emph{Proceedings of the 27th Conference on Computational Natural Language Learning (CoNLL)}, pages 294--313, Singapore. Association for Computational Linguistics.

\bibitem[{Wei et~al.(2023)Wei, Wei, Tay, Tran, Webson, Lu, Chen, Liu, Huang, Zhou, and Ma}]{wei2023larger}
Jerry Wei, Jason Wei, Yi~Tay, Dustin Tran, Albert Webson, Yifeng Lu, Xinyun Chen, Hanxiao Liu, Da~Huang, Denny Zhou, and Tengyu Ma. 2023.
\newblock \href {http://arxiv.org/abs/2303.03846} {Larger language models do in-context learning differently}.

\bibitem[{Xie et~al.(2023)Xie, Zhang, Chen, Lou, and Su}]{xie2023adaptive}
Jian Xie, Kai Zhang, Jiangjie Chen, Renze Lou, and Yu~Su. 2023.
\newblock \href {http://arxiv.org/abs/2305.13300} {Adaptive chameleon or stubborn sloth: Revealing the behavior of large language models in knowledge conflicts}.

\bibitem[{Yoo et~al.(2022)Yoo, Kim, Kim, Cho, Jo, Lee, goo Lee, and Kim}]{yoo2022groundtruth}
Kang~Min Yoo, Junyeob Kim, Hyuhng~Joon Kim, Hyunsoo Cho, Hwiyeol Jo, Sang-Woo Lee, Sang goo Lee, and Taeuk Kim. 2022.
\newblock \href {http://arxiv.org/abs/2205.12685} {Ground-truth labels matter: A deeper look into input-label demonstrations}.

\bibitem[{Zhao et~al.(2021)Zhao, Wallace, Feng, Klein, and Singh}]{pmlr-v139-zhao21c}
Zihao Zhao, Eric Wallace, Shi Feng, Dan Klein, and Sameer Singh. 2021.
\newblock \href {https://proceedings.mlr.press/v139/zhao21c.html} {Calibrate before use: Improving few-shot performance of language models}.
\newblock In \emph{Proceedings of the 38th International Conference on Machine Learning}, volume 139 of \emph{Proceedings of Machine Learning Research}, pages 12697--12706. PMLR.

\bibitem[{Zheng et~al.(2023)Zheng, Zhou, Meng, Zhou, and Huang}]{zheng2023large}
Chujie Zheng, Hao Zhou, Fandong Meng, Jie Zhou, and Minlie Huang. 2023.
\newblock Large language models are not robust multiple choice selectors.
\newblock \emph{arXiv e-prints}, pages arXiv--2309.

\bibitem[{Zhong et~al.(2023)Zhong, Cui, Guo, Liang, Lu, Wang, Saied, Chen, and Duan}]{zhong2023agieval}
Wanjun Zhong, Ruixiang Cui, Yiduo Guo, Yaobo Liang, Shuai Lu, Yanlin Wang, Amin Saied, Weizhu Chen, and Nan Duan. 2023.
\newblock Agieval: A human-centric benchmark for evaluating foundation models.
\newblock \emph{arXiv preprint arXiv:2304.06364}.

\end{thebibliography}
